\documentclass[onefignum,onetabnum]{siamonline250211}

\usepackage{amsmath, amssymb, mathrsfs}
\usepackage{lipsum}
\usepackage{amsfonts}
\usepackage{graphicx,mathtools}
\usepackage{epstopdf}
\usepackage{algorithmic}
\usepackage[titletoc]{appendix}
\usepackage{bm}
\ifpdf
  \DeclareGraphicsExtensions{.eps,.pdf,.png,.jpg}
\else
  \DeclareGraphicsExtensions{.eps}
\fi

\usepackage{enumitem}
\setlist[enumerate]{leftmargin=.5in}
\setlist[itemize]{leftmargin=.5in}

\newsiamremark{remark}{Remark}
\newsiamremark{hypothesis}{Hypothesis}
\crefname{hypothesis}{Hypothesis}{Hypotheses}
\newsiamthm{claim}{Claim}
\newsiamremark{fact}{Fact}
\crefname{fact}{Fact}{Facts}

\usepackage{subcaption}

\def\R{{\mathbb R}}
\def\OPT{{\mathrm{OPT}}}
\def\Cov{{\mathrm{Cov}}}

\def\N{{\mathbb N}}
\def\E{{\mathbb E}}
\def\P{{\mathbb P}}

\def\AA{{\mathcal A}}

\DeclareMathOperator*{\argmin}{arg\,min}

\def\XX{{\mathcal X}}
\def\H{{\mathbb H}}

\def\b{{\bm b}}

\def\iid{{\mathrm{i.i.d.}}}

\def\supp{{\mathrm{supp}}}
\def\cond{{\mathrm{cond}}}
\def\col{{\mathrm{col}}}
\def\e{{\varepsilon}}

\renewcommand{\d}[1]{\ensuremath{\operatorname{d}\!{#1}}}

\headers{Hybrid least squares}{B. Adcock and B. Hientzsch and A. Narayan and Y. Xu}

\title{Hybrid least squares for learning functions from highly noisy data\thanks{Submitted to the editors on \today.
\funding{B. Adcock is supported by NSERC through grant RGPIN-2021-611675. A. Narayan is partially supported by NSF DMS-1848508, NSF DMS-2136198, AFOSR FA9550-20-1-0338, and AFOSR FA9550-23-1-0749. Y. Xu is supported by start-up funding from the
University of Kentucky and by the AMS-Simons Travel Grant 3048116562.}}}

\author{
Ben Adcock\thanks{Department of Mathematics, Simon Fraser University (\email{adcockb@sfu.ca}).}
\and Bernhard Hientzsch\thanks{Courant Institute of Mathematical Sciences, New York University (\email{bh38@nyu.edu}).}
\and Akil Narayan\thanks{Scientific Computing and Imaging Institute, University of Utah (\email{akil@sci.utah.edu}).}
\and Yiming Xu\thanks{Department of Mathematics, University of Kentucky
  (\email{yiming.xu@uky.edu}).}}

\usepackage{amsopn}

\DeclareMathOperator{\tr}{tr}

\begin{document}

\maketitle

\begin{abstract}
Motivated by the need for efficient estimation of conditional expectations, we consider a least-squares function approximation problem with heavily polluted data. Existing methods that are effective in the small-noise regime are suboptimal when large noise is present. To address this issue, we propose a hybrid approach that combines Christoffel sampling with optimal experimental design. We show that the proposed algorithm enjoys appropriate optimality properties for both sample point generation and noise mollification, leading to improved computational efficiency and sample complexity compared to existing methods. We also extend the algorithm to convexity-constrained settings with similar theoretical guarantees. When the target function is defined as the expectation of a random field, we further extend our approach to leverage adaptive random subspaces and establish results on the approximation capacity of the adaptive procedure. Our theoretical findings are supported by numerical studies on both synthetic data and on a more challenging stochastic simulation problem in computational finance.
\end{abstract}



\section{Introduction}\label{sec:1}
Efficient computation of conditional expectations is of significant interest in stochastic computation \cite{glasserman2004monte, williams2006gaussian, xiu2010numerical}. A common task involves swiftly assessing conditional expectations across a large number of conditioning parameters. For instance, in computational finance, such a scenario arises when approximating prices of financial instruments represented as $f(x) = \E[u(S_t(x), x)\,|\, x]$ for various $x$ on a dense grid. The payoff function $u$ depends on the stochastic process $S_t$ and thus is random, and $x$ denotes the parameters of interest. When the dimension of $x$ is moderately large, classical approaches such as Monte Carlo (MC) simulation or Feynman--Kac formulaic procedures can become inefficient and computationally onerous. 

One popular alternative is to construct a \textit{surrogate model} for $f$, which would entail collecting an ensemble of realizations $\{f(x_i)\}_{i\in [m]}$ and fitting a model or response surface to this data. For example, recent machine learning techniques have been utilized to train such surrogate models \cite{huge2020differential, polala2023parametric}. These approaches are proving more computationally tractable for high-dimensional problems, owing largely to the expressive power of nonlinear approximation classes such as neural networks and the availability of modern software infrastructure for training them \cite{paszke2017automatic}. However, these approaches are predominantly empirical, can require a very large amount of data ($m$), and often require intricate hyperparameter/architectural tuning. As a result, they may be less suitable in settings in which training time is limited and rigor, trustworthiness, and certification are desired. 

In this work, we consider an alternative linear parameterization based on least-squares approximation. In particular, a sample of $u(S_t(x), x)$ can be viewed as an unbiased observation of $f(x)$ contaminated by potentially large noise. In the following, we formulate the noisy function approximation problem in a more general setting.

\subsection{Problem setup}
Let $\Omega\subset\R^d$ and $\mu\in\mathcal P(\Omega)$, where $\mathcal P(\Omega)$ denotes the set of probability measures on $\Omega$. For a function $f\in L^2_\mu(\Omega) \coloneqq\left\{g: \Omega\to\R \,|\, \int_{\Omega}g^2(x)\mu(\d x)<\infty\right\}$ and a prescribed $n$-dimensional subspace $V_n\subset L^2_\mu(\Omega)$, the least-squares problem concerns finding the orthogonal projection of $f$ in $V_n$ with respect to the norm $\|\cdot\|_{L^2_\mu}$:
\begin{align}
  f^\ast &=\argmin_{g \in V_n}\left\|f- g\right\|^2_{L^2_{\mu}}. \label{cls}
\end{align}
Given an(y) orthonormal basis $\{v_i\}_{i\in [n]}$ of $V_n$, the least-squares solution $f^*$ can be explicitly expressed through a coefficient vector,
\begin{align}
 &f^* = \argmin_{\bm\alpha \in \R^n} \left\|f- \sum_{i \in [n]} \alpha_i v_i \right\|^2_{L^2_{\mu}} =  \sum_{i\in [n]}\alpha^*_iv_i, & \text{where}  \quad \alpha_i^* = \langle f, v_i\rangle_{L^2_{\mu}},  
  \quad i\in [n].\label{alphaoracle}
\end{align}

Typically, $\bm\alpha^*=(\alpha_1^*, \ldots, \alpha_n^*)^\top$ cannot be exactly calculated due to limited information about $f$, and one often needs to discretize the problem for computation. In our setting, we assume that $f$ is unobservable directly, but instead that we can observe noisy evaluations of $f$. Our observation model $y(x)$ is given by
\begin{align}
 y(x) = f(x) + \e(x), \quad\quad x\in \Omega, \label{model_noise}
\end{align}
where $\e(x)$ is centered and uncorrelated with the sigma-field generated by $x$, i.e., 
\begin{align*}
f(x) = \E[y(x)\,|\, x],\quad\quad\sigma^2(x) = \text{Var}[y(x)\,|\, x]>0. 
\end{align*}

At this stage, we place no particular restrictions on $\sigma$, and allow $\sigma(x)/|f(x)| \gg 1$. Under this model, a general approach for discretization is based on random sampling \cite{adcock2024optimal}. This procedure first samples a set of points $\XX =\{x_i\}_{i\in [m]}$ followed by solving a discrete least-squares problem based on evaluations of $f$ on $\XX$. By taking $\iid$ samples $\XX$ from a measure $\nu\in\mathcal P(\Omega)$, where $\nu(\d x) = w^{-1}(x)\mu(\d x)$ and $w^{-1}>0$ satisfying $\int_\Omega w^{-1}(x)\mu(\d x) = 1$, and noisy observations $\{y_i\}_{i\in [m]}$ generated from \eqref{model_noise}, one can solve the following weighted least-squares problem to compute an approximate solution for \eqref{alphaoracle}:
\begin{align}
&\bm W^{\frac{1}{2}}\bm V\bm\alpha = \bm W^{\frac{1}{2}}\bm b, \quad\quad \bm b\coloneqq\frac{1}{\sqrt{m}}(y_1, \ldots, y_m)^\top,\label{myls}
\end{align}
where
\begin{align}\label{mywv}
\bm W \coloneqq \begin{bmatrix}
    w(x_1) & & \\
    & \ddots & \\
    & & w(x_m)
  \end{bmatrix}\in\R^{m\times m}, 
  \quad\bm V \coloneqq \frac{1}{\sqrt{m}}\begin{bmatrix}
    v_1(x_1) & \cdots & v_n(x_1)\\
    &  & \\
    v_1(x_m) & \cdots & v_n(x_m)
  \end{bmatrix}\in\R^{m\times n}.
\end{align}
Note that \eqref{myls} approximates \eqref{alphaoracle} by replacing the reference measure $\mu$ by a random weighted empirical measure on $\XX$ that converges weakly to $\mu$ with probability one as $m\to\infty$. This procedure is a special instance of the general framework of importance sampling-based empirical risk minimization in machine learning \cite{vapnik1991principles}.  

Denote a solution to \eqref{myls} as $\widehat{\bm\alpha}$ and $\widehat{f}$ the corresponding approximant. The accuracy of $\widehat{f}$ compared to $f^*$ was investigated in \cite{cohen2013stability} when $w\equiv 1$ using matrix concentration. Subsequent works \cite{cohen2017optimal, narayan2017christoffel} extended the idea to the case of general weights and identified $w$ that achieves the optimal sample complexity using the Christoffel function of $V_n$ \cite{nevai1986geza}:
\begin{align}
w(x) = \frac{n}{\Phi_n(x)}, \quad\quad\Phi_n(x)\coloneqq \sup_{v \in V_n: \ \|v\|_{L^2_\mu} = 1}|v(x)| = \sum_{i\in [m]}v^2_i(x).\label{chris}
\end{align}

The definition of $\Phi_n$ is independent of the choice of basis, and the corresponding sampling measure $\nu$ in the context of least squares is often called the \textit{optimal measure} or \textit{induced measure}; sampling from this measure is called \textit{Christoffel sampling}. Generating $\XX$ with Christoffel sampling results in the following approximation error bound.

\begin{theorem}\label{thm:cm17}
Under the optimal choice of $w$ in \eqref{chris}, there exists some event $\AA$ and an absolute constant $c>0$ such that if $m\gtrsim n\log n$, then $\P(\AA) = 1- n^{-2}$ and   
\begin{align}
\E\left[\left\|\widehat{f}-f\right\|^2_{L^2_\mu}\,|\, \AA\right]\leq \left(1+\frac{cn}{m}\right)\OPT + \frac{n}{m}\|\sigma\|^2_{L^2_\nu}, \quad\quad\OPT\coloneqq \left\|f-f^*\right\|^2_{L^2_\mu}.\label{oracle-l2}
\end{align}
\end{theorem}

The event $\AA$ corresponds to the randomness arising from the $m$-fold $\nu$-sampling used to generate $\XX$; the randomness of the noise in the samples $y_i$, contained in the vector $\b$, plays no role in determining $\AA$. Roughly speaking, $\AA$ contains realizations of $\XX$ on which $\bm W^{\frac{1}{2}}\bm V$ is well-conditioned; see \eqref{myAA} for a definition. The exact form of \Cref{thm:cm17} is not explicitly stated in the literature but can be deduced from existing results. For example, one can adapt the result for the conditioned weighted least-squares estimator in \cite[Theorem 4.1 (ii)]{cohen2017optimal} to the unconditioned weighted least-squares estimator with conditional expectation using Markov's inequality. This adaptation provides a bound comparable to \eqref{oracle-l2}, but with a noise dependence term of $n\|\sigma\|^2_{L^\infty_\mu}/m$.  The improved noise dependence to $n\|\sigma\|^2_{L^2_\nu}/m$ can be obtained for free by performing the same estimates above \cite[Eq.~(4.5)]{cohen2017optimal} without the last inequality. The result in \Cref{thm:cm17} is independent of the dimension $d$ and the geometries of $\Omega$ and $V_n$; the information of these objects is codified in the design of Christoffel sampling. Similar randomized least-squares methodology has found extensive applications in scientific computing and numerical approximation \cite{avron2017random, guo2018weighted, nelsen2021random, malik2021sampling, xu2023randomized}; see also \cite{guo2020constructing, hadigol2018least, adcock2022sparse, martinsson2020randomized, adcock2024optimal} for detailed results and surveys on related topics.

The result \Cref{thm:cm17} motivates the work of this paper: For any fixed $\eta\geq\OPT$, the error bound in \eqref{oracle-l2} is $\mathcal O(\eta)$ if $m\gtrsim n\max\{\log n, \|\sigma\|^2_{L^2_\nu}/\eta\}$. When $\sigma\equiv 0$, this becomes $m\gtrsim n\log n$, which matches the lower bound $n$ up to a logarithmic factor and thus is near-optimal.  When $\|\sigma\|^2_{L^2_\nu}$ is very large, $\|\sigma\|^2_{L^2_\nu}/\eta$ becomes dominant over the $\log n$ factor, making the optimality of the statement in \Cref{thm:cm17} ineffective. 

Informally, when the noise pollution is larger than the orthogonal projection error $\OPT$, one must invest extra sampling simply to resolve noise instead of approximating the function. While this seems reasonable, the procedure corresponding to \Cref{thm:cm17} generates samples $\XX$ at \textit{different} locations to resolve heterogeneous noise. Intuitively, one expects that it is more efficient to sample at $|\XX| = m \sim n \log n$ locations first to resolve the deterministic behavior of $f$, and then \textit{repeatedly} sample at locations in $\XX$ to average out noise, with a heterogeneous sample allocation to account for the different noise pollution values on $\XX$. This is precisely the high-level procedure we propose and analyze in this paper.

One branch of existing work that addresses function approximation in the large-noise setting models large noise as \textit{corruptions}, i.e., a fraction of the samples is assumed to be highly contaminated, while many samples have small or zero noise \cite{li_compressed_2012, shin_correcting_2016, adcock_compressed_2018}. In contrast, we assume a more general model in which all samples can be corrupted. Another approach is to use alternative statistical analysis to address samples polluted with spatially homogeneous, and possibly large, noise \cite{matsuda_polynomial_2024}. However, this analysis considers a particular deterministic sampling procedure in a single spatial dimension with approximation from polynomial subspaces. Our approach addresses the more general scenario when all samples can be polluted with large, heterogeneous noise in multiple spatial dimensions with a general approximation subspace.

\subsection{Contributions}
To tackle the challenges above, we propose a hybrid least-squares approach for function approximation when significant noise exists. Our contributions can be summarized as follows.
\begin{itemize}
\item We first apply (Christoffel) sampling to turn \eqref{alphaoracle} into a discrete least-squares problem. This step relies only on $(\Omega,\mu,V_n)$. The second step, which we refer to as ``function evaluation'', aims to mitigate noise introduced by $\e(x)$. Instead of taking more single evaluations over $\Omega$ with respect to $\nu$, we employ a weighted MC procedure to estimate the values of $f$ \emph{only} on the sample points $\XX$ (\Cref{alg:2}). This step is new. Fixing a total number of affordable samples $L$, the determination of where and how much to repeatedly sample on $\XX$ is an \textit{allocation} problem. The allocation can be optimized using experimental design criteria and viewed as another step of importance sampling. The combination of these two steps gives rise to the \textit{hybrid} least-squares algorithm (\Cref{alg:1}). For the proposed hybrid least-squares algorithm, we establish in \Cref{thm:main} an error bound for sample complexity and demonstrate its superiority over the standard optimally reweighted least-squares provided in \Cref{thm:cm17}. 
\item  Motivated by applications of structure-preserving noisy least-squares approximation, we extend our results to a constrained least-squares setting with additional convexity constraints. We show in \Cref{thm:conv} that the approximate least-squares solution obtained by \Cref{alg:1}, when projected onto the constraints, yields an approximate solution to the constrained problem that enjoys similar optimality guarantees. 
\item We augment our procedures by selecting $V_n$ through random adaptive subspaces. In practice, the choice of $V_n$ plays a critical role in the success of the algorithm. Even in the noiseless case, the value of $\OPT$ in \eqref{oracle-l2} can be large for poorly selected subspaces $V_n$. Although universal approximation classes, such as polynomials, are commonly used for $V_n$, they are data-oblivious and may not always be appropriate for specific tasks. When $f$ is defined as the expectation of a random field, we construct adaptive random subspaces for $V_n$ as a data-driven alternative. We establish two approximation results concerning its approximation capacity, including a law of large numbers type baseline (\Cref{thm:rb}) and a more refined analysis (\Cref{thm:984}) that showcases practical efficiency when the associated covariance kernels are approximately low-rank. Numerical simulations based on synthetic data and a more challenging stochastic simulation problem in computational finance are provided to support our theoretical findings. 
\end{itemize}

\subsection{Organization}
The rest of the paper is organized as follows. In \Cref{sec:2}, we review least squares from the perspectives of function approximation and statistical estimation, respectively, and point out their connections to our setup. In \Cref{sec:3}, we propose a hybrid least-squares framework for computing an approximate solution to \eqref{alphaoracle} based on weighted MC estimation. In \Cref{sec:opt}, we instantiate the abstract algorithm in \Cref{sec:3} with two least-squares decoders and identify the approximate optimal allocations under specific experimental design criteria. In \Cref{sec:4}, we combine the ideas in \Cref{sec:3} and \Cref{sec:opt} to obtain a practical algorithm and analyze its theoretical performance, followed by an extension to the convexity-constrained setting. In \Cref{sec:rs}, we construct adaptive random subspaces to approximate the target function $f$ for a general class of $f$ and investigate their approximation efficiency. In \Cref{sec:num}, we present numerical studies to verify our theoretical findings. Detailed proofs are provided in the appendices at the end of the article.

\subsection{Notation}\label{sec:nota}
For any $\bm z= (z_1, \ldots, z_n)^\top\in\R^n$, its $\ell_p$-norm is denoted by $\|\bm z\|_p$ for $1\leq p\leq\infty$. We use $\|\bm z\|_0$ to denote the cardinality of the support of $\bm z$, i.e., $\|\bm z\|_0 = |\supp(\bm z)|$. For matrices $\bm A, \bm B\in\R^{m\times n}$, $\|\bm A\|_2$ and $\cond(\bm A)$ represent the spectral norm and condition number of $\bm A$, respectively. The pseudoinverse of $\bm A$ is denoted by $\bm A^\dagger$, which coincides with the regular inverse $\bm A^{-1}$ when $\bm A$ is invertible. We use $\col(\bm A)$ to denote the column space of $\bm A$. When $m=n$, we use $\tr(\bm A)$ to denote the trace of $\bm A$. We use the notation $\bm A\succeq\bm B$ to denote the Loewner order on positive semi-definite matrices. 

For function approximation, we let $\Omega\subset\R^d$ denote a domain and $\mathcal P(\Omega)$ the set of probability measures on $\Omega$. As a special instance, we use $\mathcal P_m =\mathcal P([m]) = \{\bm q\in\R^m: \|\bm q\|_1 = 1, \bm q\geq 0\}$, the set of probability measures on $m$ distinct points, which is identified as the probability simplex in $\R^m$. For the sampling measure $\nu(\d x) = w^{-1}(x)\mu(\d x)$, we assume $w^{-1}(x)>0$ to ensure that $\mu$ and $\nu$ are equivalent. When taking $w(x) = \Phi_n^{-1}(x)$ as the inverse Christoffel function associated with $V_n$ in \eqref{chris}, this assumption is satisfied whenever $V_n$ contains the constant functions.

\section{Two perspectives on least squares}\label{sec:2}
While the problem discussed in \Cref{sec:1} pertains to function approximation, the inclusion of noise suggests a natural connection to the least-squares estimation in statistics. This section clarifies their connections and differences, which will guide us to design a hybrid framework in the subsequent sections.

The function approximation problem \eqref{alphaoracle} is deterministic in nature. When evaluations are noiseless, the only randomness while solving the least-squares problem \eqref{myls} arises from the Christoffel sampling procedure. This procedure aims to preserve the mutual orthogonality of the orthonormal basis $\{v_i\}_{i\in [n]}$ in $V_n$ under the discrete measure. The optimal measure \eqref{chris} in this case is a special instance of Lewis' change of density \cite{lewis1978finite} that extends to general $L^p$ subspace embedding and approximation \cite{cohen2015lp}. When $\Omega$ is a finite set and $\mu$ is the uniform measure on $\Omega$, the induced measure \eqref{chris} is equivalent to the leverage score sampling \cite{malik2022fast}, which has been extensively studied in randomized numerical linear algebra \cite{woodruff2014sketching, martinsson2020randomized, murray2023randomized}. It is worth noting that this approach relies only on the approximation space $V_n$. 

Least-squares problems in the statistics literature are often grounded in a generative model with an emphasis on the estimation and inference of model coefficients. In a classical linear regression problem with fixed design matrix $\bm X\in\R^{m\times n}$, for instance, the observation vector $\bm Y\in\R^{m}$ is assumed to be generated from a linear combination of $n$ columns of $\bm X$ contaminated by noise:
\begin{align}
\bm Y = \bm X\bm\beta + \bm\eta,\label{lm}
\end{align}
where $\bm\beta\in\R^{n}$ is the model coefficient vector and $\bm\eta\in\R^{m}$ is a centered noise vector with covariance matrix $\bm\Sigma\in\R^{m\times m}$.  
In this setup, the uncontaminated observation is within the column space of $\bm X$, i.e., $\E[\bm Y] = \bm X\bm\beta\in\col(\bm X)$. The only source of randomness comes from the noisy component $\bm\eta$. In such situations, the objective is to estimate the true parameter $\bm\beta$. The generalized Gauss--Markov theorem (also called Aitken's theorem) identifies the best linear unbiased estimator of $\bm\beta$ as the weighted least-squares solution with weights determined by a whitening transformation of the noise,
\begin{align}
\widehat{\bm\beta}\coloneqq (\bm X^\top\bm\Sigma^{-1}\bm X)^{-1}\bm X^\top\bm\Sigma^{-1}\bm Y.\label{rew}
\end{align}
That is, for any other linear unbiased estimator $\widetilde{\bm\beta}$ for $\bm\beta$, $\Cov[\widehat{\bm\beta}]\preceq\Cov[\widetilde{\bm\beta}]$ \cite{johnson2002applied}. The estimator $\widehat{\bm\beta}$ is called the best linear unbiased estimator, with mean-squared error (MSE) equal to
\begin{align}
\E\left[\|\widehat{\bm\beta}-\bm\beta\|_2^2\right] = \tr(\Cov[\widehat{\bm\beta}]) = \tr((\bm X^\top\bm\Sigma^{-1}\bm X)^{-1}).\label{msl} 
\end{align}

The discrete least-squares formulation \eqref{myls} resulting from random sampling deceptively resembles \eqref{lm} with $\bm X = \bm W^{\frac{1}{2}}\bm V$ and $\bm Y = \bm W^{\frac{1}{2}}\bm b$. However, when writing $\bm Y = \E[\bm Y] + (\bm Y-\E[\bm Y])$, the noiseless term conditional on $\XX$, $\E[\bm Y\,|\,\XX] = \E[\bm W^{\frac{1}{2}}\bm b\,|\,\XX]$ is not necessarily in $\col(\bm X)$. Note that $\E[\bm Y\,|\, \XX]\in \col(\bm X)$ only if $f-f^*$ vanishes at $x_i$, i.e., $f\in V_n$. This disparity manifests as an approximation bias, in which case the weighted estimator \eqref{rew} is no longer optimal. We will analyze this additional bias term in \Cref{sec:opt}.

\section{Hybrid least squares}\label{sec:3}

In this section, we propose a hybrid least-squares framework for solving \eqref{alphaoracle}. The proposed method consists of two steps. The first step transforms \eqref{alphaoracle} into a discrete least-squares problem using random sampling, temporarily ignoring noise. For generality, here we consider an arbitrary sampling measure $\nu(\d x) = w^{-1}(x)\mu(\d x)$, which is not necessarily the optimal measure, although we will adopt this choice later in \Cref{sec:4}. This step gives rise to the following overdetermined linear system: 
\begin{align}
\bm W^{\frac{1}{2}}\bm V\bm\alpha = \bm f,\quad\quad \bm f\coloneqq\frac{1}{\sqrt{m}}(\sqrt{w(x_1)}f(x_1), \ldots, \sqrt{w(x_m)}f(x_m))^\top ,\label{lso}
\end{align}
where $\E[\bm W^{\frac{1}{2}}\bm b] = \bm f$. A least-squares solution to \eqref{lso} can be represented as 
\begin{align}
\bar{\bm\alpha} \coloneqq  (\bm W^{\frac{1}{2}}\bm V)^\dagger\bm f.\label{alphabar}
\end{align}
We reiterate that in practice we have access only to $\bm b$ defined in \eqref{myls}, and not $\bm f$, and so $\bar{\bm\alpha}$ is a noiseless oracle. The challenge we seek to overcome is that the noisy estimator $\bm W^{\frac{1}{2}}\bm b$ may have a ``large'' covariance. To address this issue, we take an additional step to replace $\bm W^{\frac{1}{2}}\bm b$ with an alternative unbiased estimator for $\bm f$, denoted as $\bm y$, utilizing weighted MC techniques. 

Let $L$ be the total number of affordable noisy function samples, and $\bm p= (p_1, \ldots, p_m)^\top\in \mathcal P_m$ be a probability vector, with each $p_i$ representing the proportion of samples allocated to the $i$th sample point $x_i$ for MC estimation. Ignoring rounding effects, we define $L_i = p_iL$ for $i\in [m]$, indicating the number of independent samples drawn for each sample point. For each $i\in [m]$, we collect $L_i$ independent samples of $y(x_i)$, denoted as $\{y_{i, j}\}_{j\in [L_i]}$. The $i$th component of the observation vector $\bm y$ is computed as the sample average weighted by $\sqrt{w(x_i)/m}$:
\begin{align}
\bm y = \frac{1}{\sqrt{m}}(\sqrt{w(x_1)}\bar{y}_1, \ldots, \sqrt{w(x_m)}\bar{y}_m)^\top, \quad\quad\bar{y}_i = \frac{1}{L_i}\sum_{j\in [L_i]}y_{i, j}, \quad\quad i\in [m].\label{myY}
\end{align} 

Assuming independence among the samples across different $i$, the first- and second-order statistics of $\bm y$ under the model \eqref{model_noise} are
\begin{align}\label{n:design}
  \E [\bm y] = \bm f, \quad\quad 
  \Cov[\bm y] = \bm\Sigma (\bm p) = \frac{1}{L}\begin{bmatrix}
    \frac{ w(x_1)\sigma^2(x_1)}{mp_1} & & 0 \\
    & \ddots & \\
     0 & & \frac{w(x_m)\sigma^2(x_m)}{ mp_m}
  \end{bmatrix}
 .
\end{align}
Substituting $\bm f$ in \eqref{lso} with the unbiased estimator $\bm y$ yields the following overdetermined linear system:
\begin{align}
\bm W^{\frac{1}{2}}\bm V\bm\alpha = \bm y. \label{lsoo}
\end{align}
Based on \eqref{lsoo}, an estimator $\widehat{\bm\alpha}(\bm p)$ for $\bar{\bm\alpha}$ can be constructed using appropriate decoders. We have gathered all the ingredients to describe the skeleton of the hybrid least-squares algorithm.  
\begin{algorithm}[H]
 \begin{algorithmic}[1]
     \STATE{Draw $m$ i.i.d. sample points $\XX = \{x_i\}_{i\in [m]}$ from $\nu$.}
     \STATE{Choose an allocation vector $\bm p$ and compute the weighted MC estimator $\bm y$. }
     \STATE{Employ decoders to construct an estimator $\widehat{\bm\alpha}(\bm p)$ for $\bar{\bm\alpha}$.} \end{algorithmic}
\caption{A skeleton of the hybrid least-squares algorithm} 
\label{alg:2}
\end{algorithm}
The remaining task is to specify the choices of $\widehat{\bm\alpha}$ and $\bm p$. Roughly speaking, given a choice of $\widehat{\bm\alpha}$, we consider an allocation $\bm p$ as optimal if it minimizes the MSE conditional on the sample points. 
In the next section, we will address this task when $\widehat{\bm\alpha}$ is either a non-reweighted or a reweighted least-squares estimator, respectively.  

It is worth emphasizing the two layers of randomness, one arising from sampling of $\XX$ and the other from function evaluations in $\bm y$. From this point forward, we use subscripts $\XX$ and $\bm y$ to denote the randomness in sampling and function evaluation, respectively, when taking expectations. Most of the results in the subsequent sections are stated conditional on $\XX$. 

\section{Optimal allocation}\label{sec:opt}

\subsection{Non-reweighted least squares}\label{sec:3.1}
We first consider the case where $\widehat{\bm\alpha}(\bm p)$ is the non-reweighted least-squares estimator: 
\begin{align}
\widehat{\bm\alpha}(\bm p) =  (\bm W^{\frac{1}{2}}\bm V)^\dagger\bm y. \label{unw}
\end{align}
In this case, $\widehat{\bm\alpha}(\bm p)$ given $\XX$ is unbiased for $\bar{\bm\alpha}$ since 
\begin{align*}
\E_{\bm y}[\widehat{\bm\alpha}(\bm p)\,|\, \XX] = (\bm W^{\frac{1}{2}}\bm V)^\dagger\E_{\bm y}[\bm y] = (\bm W^{\frac{1}{2}}\bm V)^\dagger\bm f = \bar{\bm\alpha}.
\end{align*}
The next lemma identifies an asymptotically optimal $\bm p$. We choose to make the $V_n$-dependence of optimal allocations notationally explicit, and so will write $\bm p_n$ in what follows to emphasize this dependence.

\begin{lemma}\label{lm:a1}
Given $\XX$, let $\widehat{\bm\alpha}(\bm p)$ be the non-reweighted least-squares estimator in \eqref{unw} and assume $\bm W^{\frac{1}{2}}\bm V$ has full column rank. The allocation vector $\bm p_n^* = (p^*_{n, 1}, \ldots, p^*_{n, m})^\top\in\mathcal P_m$ defined as 
\begin{align}
p^*_{n, i} = \frac{w(x_i)\sigma(x_i)\sqrt{\Phi_n(x_i)}}{\sum_{j\in [m]}w(x_j)\sigma(x_j)\sqrt{\Phi_n(x_j)}},\quad\quad i\in [m],\label{neyman}
\end{align}
is a $\cond(\bm V^\top\bm W\bm V)^2$-approximate solution to the following optimization problem:
\begin{align}
\min_{\bm p \in \mathcal P_m}\E_{\bm y}[\|\widehat{\bm\alpha}(\bm p)-\bar{\bm\alpha}\|_2^2\,|\, \XX],\label{lsjs}
\end{align}
where $\Phi_n$ is defined in \eqref{chris}. 
That is, 
\begin{align}
\E_{\bm y}[\|\widehat{\bm\alpha}(\bm p_n^*)-\bar{\bm\alpha}\|_2^2\,|\, \XX]\leq \cond(\bm V^\top\bm W\bm V)^2\cdot\min_{\bm p \in \mathcal P_m}\E_{\bm y}[\|\widehat{\bm\alpha}(\bm p)-\bar{\bm\alpha}\|_2^2\,|\, \XX].\label{multt}
\end{align}
\end{lemma}
The proof is based on a direct matrix calculation and is provided in \Cref{appd:1}. The allocation vector $\bm p_n^*$ is the same as the Neyman allocation for the strata variance sequence $\{w^2(x_i)\sigma^2(x_i)\Phi_n(x_i)\}_{i\in [m]}$. Note that $\bm p^*_n$ is a function of the sample points $\mathcal X$. To understand the asymptotic behavior of $\bm p^*_n$, we let $m\to\infty$. 
\begin{lemma}\label{nygle}
As $m\to\infty$, $\bm p_n^*$ converges weakly to $p^*$ for some $p^*\in\mathcal P(\Omega)$ $\mu$-a.s., where 
\begin{align*}
\frac{\d p^*}{\d\mu} = \frac{\sigma(x)\sqrt{\Phi_n(x)}}{\int_\Omega \sigma(z)\sqrt{\Phi_n(z)} \mu(\d z)}.
\end{align*}
\end{lemma}
\begin{proof}
For any bounded and continuous function $h: \Omega\to\R$, since $x_i$ are i.i.d. samples from $\nu$, it follows from the law of large numbers that $\nu$-a.s., 
\begin{align*}
\int_\Omega h(x){\bm p_n^*(\d x)} &=  \sum_{i\in [m]}\frac{w(x_i)\sigma(x_i)\sqrt{\Phi_n(x_i)}h(x_i)}{\sum_{j\in [m]}w(x_j)\sigma(x_j)\sqrt{\Phi_n(x_j)}}\\
& \xrightarrow{m\to\infty} \frac{\int_\Omega w(x)\sigma(x)\sqrt{\Phi_n(x)}h(x)\nu(\d x)}{\int_{\Omega}w(x)\sigma(x)\sqrt{\Phi_n(x)}\nu(\d x)} \stackrel{\d\nu = w^{-1}\d\mu}{=} \int_\Omega h(x) p^*(\d x),
\end{align*}
showing that $\bm p_n^*$ converges to $p^*$ weakly. Noting that $\mu$ and $\nu$ are equivalent (as assumed in \Cref{sec:nota}) allows us to transfer $\nu$-a.s. to $\mu$-a.s. 
\end{proof}
\begin{remark}\label{rem:independence}
The limit optimal allocation measure $p^*$ is independent of the sampling weight $w$ chosen in the function approximation step. This decoupling occurs because the weight $w$ in the allocation formula \eqref{neyman} cancels the reciprocal density $w^{-1}$ from the measure $\nu$ used to sample the points $x_i$ (as shown in the proof of \Cref{nygle}). When $\sigma\equiv\sigma_0>0$ is a constant function, $\d p^*\propto\sqrt{\Phi_n}\d\mu$, which depends solely on $V_n$ and differs from the optimal Christoffel sampling measure $\d\nu \propto \Phi_n \mathrm{d}\mu$. This suggests that the samplings required for minimizing estimation variance and projection bias are fundamentally different.
\end{remark}

\subsection{Reweighted least squares}\label{ssec:reweighted-ls}
Alternatively, one may consider $\widehat{\bm\alpha}(\bm p)$ constructed as a reweighted least-squares solution to \eqref{lsoo} using some weight matrix $\bm\Gamma\in\R^{m\times m}$: 
\begin{align}
\bm\Gamma\bm W^{\frac{1}{2}}\bm V\bm\alpha = \bm\Gamma\bm y, 
\end{align}
which has a least-squares solution
\begin{align}
\widehat{\bm\alpha}(\bm p) = (\bm\Gamma\bm W^{\frac{1}{2}}\bm V)^\dagger\bm\Gamma\bm y.\label{sdr}
\end{align}

In contrast to $\bm W$, the weight matrix $\bm\Gamma$ is introduced to rebalance the estimation variance rather than reduce the approximation bias. As discussed in \Cref{sec:2}, the estimator $\widehat{\bm\alpha}(\bm p)$ is unbiased for $\bar{\bm\alpha}$ if $\bm f\in\col(\bm W^{\frac{1}{2}}\bm V)$, with the optimal reweight matrix given by $\bm\Gamma = \bm\Sigma(\bm p)^{-\frac{1}{2}}$, where $\bm\Sigma(\bm p)$ is defined in \eqref{n:design}. However, such a statement no longer holds when $\bm f\notin\col(\bm W^{\frac{1}{2}}\bm V)$ due to the additional bias term resulting from reweighting. Under such circumstances, finding the optimal weight matrix is difficult.
Nonetheless, if $\bm f$ can be well approximated by $\col(\bm W^{\frac{1}{2}}\bm V)$, then we expect $\bm\Sigma(\bm p)^{-\frac{1}{2}}$ to provide a reasonable choice with appropriate adjustments. 

In the following discussion, we take $\bm\Gamma = \bm\Sigma(\bm p)^{-\frac{1}{2}}$ in \eqref{sdr}, and decompose $\bm f$ as 
\begin{align*}
\bm f = \bm z_1 + \bm z_2\in \col(\bm W^{\frac{1}{2}}\bm V)\oplus \col(\bm W^{\frac{1}{2}}\bm V)^\perp,
\end{align*}
where  
\begin{align}
\bm z_1 = \bm W^{\frac{1}{2}}\bm V(\bm W^{\frac{1}{2}}\bm V)^\dagger\bm f, \quad\quad \bm z_2 = (\bm I_m - \bm W^{\frac{1}{2}}\bm V(\bm W^{\frac{1}{2}}\bm V)^\dagger)\bm f.\label{myz12}
\end{align}
For $\delta\in (0, \frac{1}{m}]$, we define the regularized feasible set $\mathcal P_{m}(\delta)\coloneqq \left\{\bm q\in\R^m: \|\bm q\|_1 = 1, \bm q\geq \delta\right\}$. This regularized feasible set excludes solutions that have zero allocation on the support points, which may cause convergence issues when bias exists (see \Cref{rem:reg}). Moreover, we define the trace of the covariance of the estimator in \eqref{sdr} as 
\begin{align}
 H(\bm p) \coloneqq\tr(\bm U(\bm p)^{-1}),\quad\quad \bm U(\bm p) \coloneqq \bm V^\top\bm W^{\frac{1}{2}}\bm\Sigma(\bm p)^{-1}\bm W^{\frac{1}{2}}\bm V.\label{H-conv}
\end{align}
When $\E_{\bm y}[\bm y\,|\, \XX]\in\col(\bm W^{\frac{1}{2}}\bm V)$, $H(\bm p)$ is equal to the $\XX$-conditional MSE of \eqref{sdr}. In general, $H(\bm p)$ provides a lower bound on the $\XX$-conditional MSE of $\widehat{\bm\alpha}(\bm p)$ due to the additional bias. The next two lemmas show that $H(\bm p)$ is a convex function of $\bm p$ on $\mathcal P_{m}(\delta)$ and admits a minimizer $\bm q_n^{*}$, which provides an approximate optimal allocation for the estimator in \eqref{sdr} restricted to the feasible set $\mathcal P_m(\delta)$. Their proofs are given in \Cref{app:conxh} and \Cref{app:conxhl}. 

\begin{lemma}\label{mkjh}
Fixing $\XX$, consider the optimization problem:
\begin{align}
  \bm q_n^\ast = (q^\ast_{n,1}, \ldots, q^\ast_{n,m})^\top \in \argmin_{\bm p\in\mathcal P_m(\delta)}H(\bm p).\label{H-conv+}
\end{align}
Assume that $\bm W^{\frac{1}{2}}\bm V$ has full column rank. Then for every $\delta\in [0, \frac{1}{m}]$ and $m\geq n$, $\mu$-a.s., \eqref{H-conv+} is a convex optimization problem with a finite optimal solution $\bm q_n^\ast$ satisfying
\begin{align}
|\supp_{\delta}(\bm q_n^{*})|\leq \frac{n^2+n}{2},\quad\quad \supp_{\delta}(\bm q_n^{*})\coloneqq \{i\in [m]: q_{n,i}^*>\delta\}.\label{spt}
\end{align} 
\end{lemma} 
The objective in \eqref{H-conv} is similar to the $A$-optimality criteria in experimental design \cite{pukelsheim2006optimal}. When $\delta = 0$, there exists an optimal solution that is at most $(n^2+n)/2$-sparse. Since $(n^2+n)/2$ is independent of $m$, only a fixed number of the sample points will be used for function evaluation as $m\to\infty$. This is not a problem when $\bm f\in\col(\bm W^{\frac{1}{2}}\bm V)$ since perfect evaluations of any $n$ distinct points will result in exact recovery of $\bar{\bm\alpha}$ (assuming unisolvency). However, it may cause convergence issues otherwise.

\begin{lemma}\label{lm:a2}
Let $\delta\in (0, \frac{1}{m}]$ and $\widehat{\bm\alpha}(\bm p)$ be the reweighted least-squares estimator in \eqref{sdr} with weight matrix $\bm\Gamma = \bm\Sigma(\bm p)^{-\frac{1}{2}}$.  If we denote a solution to \eqref{H-conv+} as $\bm q_n^{*}$ and assume $w>0$ on $\Omega$, then, 
\begin{align}
\E_{\bm y}[\|\widehat{\bm\alpha}(\bm q_n^{*})-\bar{\bm\alpha}\|_2^2\,|\, \XX]\leq\frac{J_n}{\delta}\|(\bm V^\top\bm W\bm V)^{-1}\|_2\|\bm z_2\|_2^2+\min_{\bm p\in\mathcal P_m(\delta)}\E_{\bm y}[\|\widehat{\bm\alpha}(\bm p)-\bar{\bm\alpha}\|_2^2\,|\, \XX],\label{123511}
\end{align}
where $\bm z_2$ is defined in \eqref{myz12} and $J_n$ is the condition number of $w(x)\sigma^2(x)$ on $\Omega$ defined as $J_n = \sup_{x\in\Omega}|w(x)\sigma^2(x)|/\inf_{x\in\Omega}|w(x)\sigma^2(x)|$. 
\end{lemma}

\begin{remark}\label{rem:reg}
  The regularization parameter $\delta$ ensures that the reweighting matrix $\bm\Sigma(\bm p)^{-\frac{1}{2}}$ is non-singular. This leads to the $1/\delta$ factor the first term in the upper bound in \eqref{123511} and thus ensures that it remains bounded. Generally, the bound in \eqref{123511} is useful when $\|\bm z_2\|_2$ is small. This occurs when $V_n$ well approximates $f$, i.e., when $\OPT$ in \Cref{thm:cm17} is sufficiently small. For instance, this happens if $f$ is the expectation of some random field with a low-rank covariance function and $V_n$ is chosen as a subspace spanned by random realizations of the random field (see \Cref{sec:rs}). When $\bm\|\bm z_2\|_2 = 0$, taking $\delta\to 0$ recovers the result in \eqref{msl}.  An extended discussion is given after \Cref{thm:main}.  
\end{remark}

\begin{remark}
The approximate optimality bound in \eqref{123511} is additive rather than multiplicative as in \eqref{multt} due to the reweighting bias. Moreover, in contrast to the non-reweighted case (\Cref{rem:independence}), the additive error has an explicit dependence on the choice of weight $w$. 
\end{remark}

\section{Hybrid least-squares algorithms and error bounds}\label{sec:4}

In this section, we first combine the results in \Cref{sec:3} and \Cref{sec:opt} to obtain a hybrid algorithm for solving \eqref{alphaoracle}.  
Then we discuss an extension of the approach to tackling noisy function approximation problems with additional convexity constraints. 

\subsection{Unconstrained function approximation}

The hybrid least-squares algorithm for the non-reweighted and reweighted least-squares estimators with optimal sampling and allocation is characterized in \Cref{alg:1}. It contains the essential ingredients described in previous sections, and affords the option of choosing either the non-reweighted least squares procedure of \Cref{sec:3.1}, or the reweighted one of \Cref{ssec:reweighted-ls}. In addition, when $\sigma$ is unknown, it provides an empirical procedure through a pilot parameter $R$ that estimates $\sigma$ on $\XX$.

\begin{algorithm}
\hspace*{\algorithmicindent} \begin{tabbing}
\textbf{Input}: \= a reference measure $\mu$; \\
\> a target function evaluator $y(x)$; \\
\> an orthonormal basis $\{v_i\}_{i\in [n]}$ of $V_n$; \\
\> the conditional variance function $\sigma^2(x)$ (alternative); \\
\> the sample points size $m \geq n$; \\
\> the total evaluation sample size $L \coloneqq \gamma m$, where $\gamma \geq 1$; \\
\> the regularization parameter $\delta > 0$; \\
\> the pilot variance estimation parameter $R$.\\
\textbf{Output}: an estimate $\widehat{f}$ for $f^*\coloneqq\argmin_{v\in V}\|f-v\|^2_{L^2_\mu}.$
\end{tabbing}    
\begin{algorithmic}[1]
   \STATE Compute the induced measure $\nu$ associated with the reciprocal Christoffel function $w(x)$:
   \begin{align*}
   &d\nu = w(x)^{-1} d\mu& w(x) = \frac{n}{\Phi_n(x)} = \frac{n}{\sum_{i\in [n]}v_i^2(x)}.
   \end{align*}
     \STATE{Draw $m$ i.i.d. sample points $\XX = \{x_i\}_{i\in [m]}$ from $\nu$.}
     \IF{$\sigma^2(x)$ is \textit{not} given}{
     \STATE Estimate the conditional variance function $\sigma^2(x)$ on $\XX$ using MC with $Rm$ samples.
     }
     \ENDIF
     \STATE{Compute the allocation vector $\bm p_n$ on $\XX$:
     \begin{itemize}
     \item (Non-reweighted least-squares) Compute $\bm p_n$ as $\bm p^*_n$ in \eqref{neyman};
     \item (Reweighted least-squares) Compute $\bm p_n$ as an optimal solution $\bm q^{*}_n$ to \eqref{H-conv+}.
     \end{itemize}}
     \STATE{Compute the evaluation vector $\bm y$ using \eqref{myY} with total sample size $L$ and allocation $\bm p_n$.}
     \IF{$\bm p_n=\bm p^*_n$}{
     \STATE Solve the non-reweighted least-squares $\bm W^{\frac{1}{2}}\bm V\bm\alpha = \bm y$ where $\bm V, \bm W$ are defined in \eqref{mywv}:
     \begin{align*}
     \widehat{\bm \alpha} = (\bm W^{\frac{1}{2}}\bm V)^\dagger\bm y.
     \end{align*}}
     \ELSIF{$\bm p_n=\bm q^{*}_n$}{
     \STATE Solve the reweighted least-squares $\bm\Sigma(\bm q^{*}_n)^{-1/2}\bm W^{\frac{1}{2}}\bm V\bm\alpha = \bm\Sigma(\bm q^{*}_n)^{-1/2}\bm y$ where $\bm\Sigma(\bm q^{*}_n)$ is defined in \eqref{n:design}: 
     \begin{align*}
     \widehat{\bm \alpha} = \left(\bm\Sigma(\bm q^{*}_n)^{-1/2}\bm W^{\frac{1}{2}}\bm V\right)^\dagger\bm\Sigma(\bm q^{*}_n)^{-1/2}\bm y.
     \end{align*}}
     \ENDIF
 \STATE{Compute $\widehat{f}$ as $\widehat{f} = \sum_{i\in [m]}\widehat{\alpha}_iv_i$.}
 \end{algorithmic}
\caption{Hybrid least-squares algorithms with optimal allocation} 
\label{alg:1}
\end{algorithm}

Our main theoretical result for \Cref{alg:1} is as follows.
\begin{theorem}[Error for \Cref{alg:1} with known $\sigma$]\label{thm:main}
Assume that $\sigma^2(x)>0$ is given and let $\OPT$ be the oracle approximation error defined in \eqref{oracle-l2}. 
Let $\Lambda(\cdot)$ denote the spectrum of a matrix and define the $\XX$-measurable event $\AA$ as 
\begin{align}
\AA=\left\{\XX: \Lambda(\bm W^{\frac{1}{2}}\bm V)\subseteq[0.9, 1.1]\right\}.\label{myAA}
\end{align}
If $m\gtrsim n\log n$, then $\P_{\XX}(\AA)>1-n^{-2}$, and there exists an absolute constant $c>0$ such that the following conditional error bounds hold for the output of \Cref{alg:1}:  
\begin{enumerate}
\item If $\bm p_n = \bm p_n^*$, then
\begin{align}
\E_{\XX, \bm y}\left[\left\|\widehat{f}-f\right\|^2_{L^2_\mu}\,|\,  \AA\right]&\lesssim \OPT + \E_{\XX}[G(\bm p^*_n)]\label{p1-1},
\end{align}
where $G(\bm p) \coloneqq \frac{1}{L}\sum_{i\in [m]}\frac{w^2(x_i)\sigma^2(x_i)\Phi_n(x_i)}{m^2p_i}$ and 
\begin{align}
\E_{\XX}[G(\bm p^*_n)] &= \frac{n}{L}\left[\frac{1}{m}\|\sigma\|^2_{L^2_\mu} + (1-1/m)\left\|\sigma\sqrt{\frac{\Phi_n}{n}}\right\|^2_{L^1_\mu}\right].\label{tired}
\end{align}
\item If $\bm p_n = \bm q_n^{*}$, then
\begin{align}
\E_{\XX, \bm y}\left[\left\|\widehat{f}-f\right\|^2_{L^2_\mu}\,|\,  \AA\right]&\lesssim \frac{J_n}{\delta}\cdot\OPT + \E_{\XX}[H(\bm q_n^{*})],\label{p1-2}
\end{align}
where $J_n$ and $H$ are defined in \Cref{lm:a2} and \eqref{H-conv}, respectively. 
\end{enumerate} 
\end{theorem}
The proof is based on the existing results on Christoffel sampling and the estimates in \Cref{sec:opt}, and is provided in \Cref{app:main}. Below, we discuss the computational complexity of \Cref{alg:1} and the error bounds in \Cref{thm:main}.

\paragraph{Computational complexity}
The Christoffel sampling procedure in steps 1-2 of \Cref{alg:1} is random and often computationally cheap compared to the subsequent function evaluation. To reduce randomness and improve accuracy, one may consider an additional boosting procedure over several random samplings to further improve accuracy \cite{haberstich2022boosted}. 

For function evaluation, in addition to collecting $L = \gamma m$ samples, one must also determine the allocation weights. For reweighted least squares, this requires solving a constrained optimization problem of dimension $m$. The gradient of the objective function can be explicitly computed in $\mathcal{O}(mn^2)$ operations. By using a quasi-Newton method, such as Sequential Least Squares Quadratic Programming (SLSQP), the total cost per iteration is $\mathcal{O}(mn^2 + m^3)$. In the regime where $m = \Theta(n \log n)$, this cost is effectively cubic in $n$, up to logarithmic factors.

For estimation, one needs to solve a least-squares problem of size $m\times n$, which has complexity $\mathcal O(mn^2)$ if direct methods are used. In contrast, under the same evaluation complexity, the standard randomized least-squares approach requires solving a least-squares problem of size $L\times n$, which has complexity $\mathcal O(Ln^2)$. This number is much larger than $mn^2$ if $\gamma$ is large.  

The analysis here assumes that the variance $\sigma^2(x)$ is given. If not, an additional evaluation budget of size $Rm$ is required to estimate $\sigma^2(x)$. Under suitable moment conditions, the cost of this pilot estimation can be quantified compared to the total evaluation cost using a perturbation analysis. A rigorous analysis is provided in \Cref{app:cost}.

\paragraph{Error bounds}
The two terms in the error bounds \eqref{p1-1} and \eqref{p1-2} correspond to the approximation error and statistical error, respectively. We begin by interpreting the bound in \eqref{p1-1}. The bounds in \eqref{p1-1} and \eqref{tired} together imply that to achieve an average error of order $\eta$ for some $\eta\geq\OPT$, one needs the total evaluation complexity
\begin{align}
L = \gamma m\gtrsim n\left[\log n+\frac{1}{\eta}\left(\frac{1}{m}\|\sigma\|^2_{L^2_\mu} + (1-1/m)\left\|\sigma\sqrt{\frac{\Phi_n}{n}}\right\|^2_{L^1_\mu}\right)\right].\label{newb}
\end{align}
Noting $\|\sigma\|^2_{L^2_\nu} = \|\sigma\sqrt{\Phi_n/n}\|^2_{L^2_\mu}$, 
\begin{align}
\frac{\frac{1}{m}\|\sigma\|^2_{L^2_\mu} + (1-1/m)\left\|\sigma\sqrt{\frac{\Phi_n}{n}}\right\|^2_{L^1_\mu}}{\left\|\sigma\right\|^2_{L^2_\nu}}= \frac{1}{m}\frac{\|\sigma\|^2_{L^2_\mu}}{\|\sigma\|^2_{L^2_\nu}} + (1-1/m)\frac{\left\|\sigma\sqrt{\Phi_n}\right\|^2_{L^1_\mu}}{\left\|\sigma\sqrt{\Phi_n}\right\|^2_{L^2_\mu}}. \label{jgutis}
\end{align} 
When $\Omega$ is compact and $\sigma(x)$ is positive and continuous, the ratio $\|\sigma\|^2_{L^2_\mu}/\|\sigma\|^2_{L^2_\nu}$ is bounded by the condition number of $\sigma^2$ on $\Omega$, which is independent of $m$. Consequently, \eqref{jgutis} approaches $\|\sigma\sqrt{\Phi_n}\|^2_{L^1_\mu}/\|\sigma\sqrt{\Phi_n}\|^2_{L^2_\mu}\leq 1$ as $m \to \infty$ by Jensen's inequality. Hence, the sample complexity in \eqref{newb} is asymptotically at least as efficient as the bound in \Cref{thm:cm17}.

The interpretation of \eqref{p1-2} is less straightforward. Compared to \eqref{p1-1}, it improves the error dependence on the estimation variance at the cost of introducing a multiplicative constant in front of the approximation bias. Setting $\delta \leq 1/(m\sqrt{J_n})$, $\bm p_n^*\in\mathcal P_m(\delta)$ so that $H(\bm q^{*}_n)\leq H(\bm p^*_n)\leq G(\bm p^*_n)$ (the first inequality follows from the definition of $\bm q_n^*$, and the second from the generalized Gauss--Markov theorem), which implies $\E_\XX[H(\bm q^{*}_n)]\leq \E_\XX[G(\bm p^*_n)]$. Meanwhile, the constant in front of $\OPT$ is lower bounded by $mJ^{3/2}_n$. Whether the gain outweighs the loss depends on the magnitude of $\OPT$. If $\OPT$ is sufficiently small, then there is an advantage. Nevertheless, a rigorous analysis under such circumstances is beyond the scope of this paper.

\subsection{Constrained function approximation}\label{sec:convls}

In some practical scenarios, least-squares approximation is carried out under additional constraints. For instance, structure-preserving approximations often impose positivity or monotonicity on the resulting approximant \cite{zala2020structure}. Similar constraints are prevalent in computational finance, where target functions, such as pricers of financial instruments with non-negative payoffs, must remain positive. Many of these requirements can be encoded as membership in a convex set. We show in this section that, with a proper redefinition of optimality, using \Cref{alg:1} to first compute an unconstrained approximation and subsequently projecting it onto the convex set results in essentially the same error bounds as for the unconstrained case.

Let $\mathcal C\subseteq V_n$ be a closed convex set. Consider the following constrained version of \eqref{cls}:
\begin{align}
&\min_{v\in\mathcal C}\left\|f- v\right\|^2_{L^2_{\mu}}.\label{convls}
\end{align} 
Note that \eqref{convls} has a unique solution. To see this, let $\Pi_{\mathcal C}: L^2_\mu(\Omega)\to \mathcal C$ be the projection operator in the distance induced by the $\|\cdot\|_{L^2_\mu}$ norm, i.e., $\Pi_{\mathcal C}(g) = \argmin_{v\in\mathcal C}\|g-v\|^2_{L^2_\mu}$ for $g\in L^2_\mu(\Omega)$, which is well-defined due to the closedness and convexity of $\mathcal C$. Denote the solution to \eqref{convls} as $f^*_c$. It follows from the Pythagorean theorem that
\begin{align}
f^*_c = \Pi_{\mathcal C}(f) = \Pi_{\mathcal C}(f^*),\label{obs}
\end{align}
where $f^*$ is the minimizer to the unconstrained problem \eqref{cls}. 
Note that $\Pi_{\mathcal C}$ is a contraction with respect to $\|\cdot\|_{L^2_\mu}$. This is well known in the convex analysis literature \cite{borwein2006convex} and we state it as the following lemma without proof. 
\begin{lemma}\label{cont}
The projection operator $\Pi_{\mathcal C}: L^2_\mu(\Omega)\to \mathcal C$ is a contraction in $\|\cdot\|_{L^2_\mu}$.
\end{lemma}
Thus, given an approximate solution to \eqref{cls}, one can compute an approximate solution to \eqref{convls} by projecting it to $\mathcal C$. 
The quality of such an approximate solution is quantified in the following theorem, whose proof can be found in \Cref{app:conv}.  
\begin{theorem}\label{thm:conv}
Let $f^*_c$ be the solution to the constrained approximation problem \eqref{convls} and let $\OPT_c = \|f-f^*_c\|^2_{L^2_{\mu}}$. 
Denote $\widehat{f}$ the approximate unconstrained solution computed by Algorithm \ref{alg:1} and $\widehat{f}_c = \Pi_{\mathcal C}(\widehat{f})$. 
Under the same assumptions in \Cref{thm:main}, the bounds in \eqref{p1-1} and \eqref{p1-2} hold with $\OPT$ and $\widehat{f}$ replaced by $\OPT_c$ and $\widehat{f}_c$, respectively, subject to an enlargement of the implicit constants by a factor of at most 2. 
\end{theorem}

\section{Random subspaces}\label{sec:rs}
Identification of an appropriate approximation subspace $V_n$ is crucial for the success of hybrid least-squares methods, especially when biased estimators are used; i.e., the term $\OPT$ in \eqref{oracle-l2} should be small. This section addresses one strategy to identify such a subspace in a data-dependent way for a special class of functions $f$ commonly arising in stochastic simulation. In this situation, we assume a more special model for $f$: the target function $f$ can be written as the expectation of some random field $g(x,Z)$: 
\begin{align}
&f(x) = \E_Z[g(x,Z)],\quad\quad g: \Omega\times\mathbb H\to\R,
\end{align}
where $g$ is a measurable function and $\H$ is the sample space for $Z$. Note that $g$ in this model can be cast in the form \eqref{model_noise} by taking $y(x) = g(x,Z)$, and $\e(x) = g(x,Z) - f(x)$. When we have access to evaluations of $g$, these can be used to identify a good candidate for $V_n$. In particular, one may consider $V_n$ spanned by random basis functions defined as follows. Given i.i.d. copies of $Z$ denoted by $Z_1, \ldots, Z_n$, define
\begin{align}
V_n = \text{span}\left\{g_i\coloneqq g(\cdot,Z_i)\right\}_{i\in [n]}\subset L^2_\mu(\Omega).\label{myVn}
\end{align}
Since the sample average of $g_i$ is the size-$n$ MC estimate of $f$, a simple dimension-free result on the approximation error of $f$ under $V_n$ can be obtained using the law of large numbers, assuming $g$ is an $L^2$ stochastic process.
\begin{theorem}\label{thm:rb}
Under the above assumptions on $f$ and with $V_n$ defined in \eqref{myVn}, then given $\e, \delta>0$, if $k = \lceil2\e^{-2}\|\sigma\|^2_{L^2_\mu}\rceil$ and $n> 1.5\log(1/\delta)k$, where $\sigma^2(x) = \mathrm{Var}[g(x, Z)\,|\, x]$, then with probability at least $1-\delta$,
\begin{align}
\min_{v\in V_n}\|f-v\|_{L^2_\mu}\leq\min_{v\in\bar{V}_{n,k}}\left\|f-v\right\|_{L^2_\mu}<\e,\label{simp}
\end{align}
where $\bar{V}_{n,k}$ is the set of linear combinations of $\{g_i\}_{i\in [n]}$ with support size no greater than $k$:
\begin{align}
\bar{V}_{n, k} \coloneqq \left\{\sum_{i\in [n]}\alpha_ig_i:\|\bm\alpha\|_0\leq k\right\}\subset V_n. \label{myconstr}
\end{align} 
\end{theorem}

\Cref{thm:rb} provides a baseline on the approximation capacity of $V_n$, similar to \cite{rahimi2008uniform}, and its proof is given in \Cref{app:mc1}. Theoretically, the least-squares approximation using the random subspaces $V_n$ in \eqref{myVn} is provably no worse than MC averaging. In certain applications, fully exploiting the approximation capacity of $V_n$ through linear coefficient optimization is essential. For example, in model calibration in computational finance, memory limitations and evaluation overhead can severely restrict the feasible sample size $n$. Indeed, each query to the surrogate requires evaluating all basis functions, while in practice, the system may store only random seeds to reconstruct the basis functions on the fly rather than maintaining a dense grid representation. Under such circumstances, it becomes crucial to fully utilize the representational power of $V_n$.

In practice, the least-squares approximation using $V_n$ may achieve much smaller approximation error than is shown in \Cref{thm:rb}. For instance, when $\Omega$ is a finite set, choosing $n=|\Omega|$ is sufficient to exactly represent $f$ provided that the $n$ random functions $g_i$ are linearly independent. This observation can be generalized by leveraging the low-rank structure of the kernel associated with $g(x; Z)$. In the following, we assume that $\Omega$ is compact. 

Let $K(x, y) = \E_Z[g(x; Z)g(y; Z)]-f(x)f(y)$ denote the covariance function of the random field $\{g(x; Z)\}_{x\in\Omega}$ and assume that $K(x, y)$ is continuous on $\Omega\times\Omega$. By Mercer's theorem, there exist an orthonormal basis $\{\phi_i\}_{i\in\N}$ in $L^2_\mu(\Omega)$ and a nonincreasing nonnegative sequence $\{\lambda_i\}_{i\in\N}\in\ell_1(\N)$ such that $K(x, y) = \sum_{i\in\N}\lambda_i\phi_i(x)\phi_i(y)$. 
The random field $g(x; Z)$ can be represented using the Karhunen--Lo\`eve (KL) expansion as
\begin{align}
&g(x; Z) = f(x) + \sum_{i\in\N}\sqrt{\lambda_i}\xi_i \phi_i(x), & \xi_i = \frac{1}{\sqrt{\lambda_i}}\int_\Omega (g(x; Z)-f(x))\phi_i(x) \mu(\d x),\label{here}
\end{align} 
where $\xi_i$'s are centered and uncorrelated random variables with unit variance. The next result shows that under suitable tail-decay conditions (which can be generalized), an effective approximation of $f$ can be achieved with $V_n$ if $K(x, y)$ is close to being of finite-rank. The proof of this result is given in \Cref{app:kernel}.

\begin{definition}[Uniformly subgaussian sequence]
A sequence of random variables $\{X_i\}_{i\in\N}$ is called uniformly subgaussian if there exists an absolute constant $c>0$ such that
\begin{align}
\sup_{i\in\N}\P(|X_i|>x)\leq 2e^{-x^2/c}\quad\quad x\geq 0. \label{subexp}
\end{align} 
\end{definition}

\begin{theorem}\label{thm:984}
 Assume that $\{\xi_i\}_{i\in\N}$ in \eqref{here} have no atoms and are uniformly subgaussian in the sense of \eqref{subexp} and let $\tau_s = \sum_{i\geq s}\lambda_i$.
Fixing $r\in\N$, there exist constant $C_1>0$ depending on $c$ only such that, with $k = \lceil C_1 r(\log r)^3\rceil$ and for any $\delta>0$, if $n> 10\log (1/\delta) k$, then with probability at least $1-\delta$,
\begin{align}
\min_{v\in V_n}\|f-v\|_{L^2_\mu}\leq \min_{v\in\bar{V}_{n,k}}\left\|f-v\right\|_{L^2_\mu}\leq 24\sqrt{r\tau_{r+1}},\label{lopbdd}
\end{align}
where $\bar{V}_{n,k}$ is the same as defined in \eqref{myconstr}.
\end{theorem}
\begin{remark}
The non-atomic assumption is not essential but simplifies the statement. Compared to \Cref{thm:rb}, the error bound in \eqref{lopbdd} depends on the decay of $\{\lambda_i\}_{i\in\N}$ rather than its $\ell_1$-norm, i.e., $\|\sigma\|^2_{L^2_\mu} = \sum_{i\in\N}\lambda_i$. This result shares similar flavors with other results in low-rank approximation \cite{alla2019randomized, peherstorfer2022breaking} but its proof involves a distinct technical treatment, particularly compared to \cite{reiss2020nonasymptotic}, which is based on analyzing empirical spectral projectors. Additionally, our approach is different from the one in kernel feature expansion \cite{bach2017equivalence}, which requires additional regularization and importance sampling on $Z$ that is not practical in our setting.
\end{remark}

In many applications, we have $\H = \R^s$ for some $s\in\N$. In this case, with fixed $Z$, $g(x; Z)$ can be evaluated as a function of $x$. To sample from the reciprocal Christoffel density in $V_n$, we adopt the strategy in \cite{adcock2020near} that consists of three steps:
\begin{enumerate}
\item Discretize the measure $\mu$ using the empirical measure of $Q$ points independently sampled from $\mu$;
\item Compute an orthonormal basis with respect to the discrete measure using QR decomposition;
\item Draw samples from the discrete reciprocal Christoffel density.
\end{enumerate}

The grid discretization error in step~1 was analyzed in \cite{adcock2020near} using a Nikolskii-type inequality. For large $d$, maintaining sufficient accuracy may require a large value of $Q$, potentially reintroducing the curse of dimensionality. This issue can be mitigated, for example, when $\Omega$ is a product domain and $\mu$ is the uniform measure, by employing more efficient discretization schemes such as low-discrepancy sequences \cite{niederreiter1992random}. Alternatively, one can avoid this bottleneck via iterative row-sampling techniques for approximate Christoffel sampling \cite{herremans2025refinement, cohen2015uniform}. These approaches typically start from a coarse sampler that is iteratively refined to achieve approximate Christoffel sampling without requiring an initial orthonormal basis.

\section{Numerical simulation}\label{sec:num}

In this section, we apply hybrid least squares to a synthetic multivariate function approximation setup and a stochastic simulation problem in computational finance. We compare the proposed algorithms with two other methods, including a naive hybrid least-squares procedure with equal allocation (i.e., $L$ is allocated equally to each sample point for MC estimation), and the other based on empirical risk minimization with training data sampled from $\mu$. The details of the algorithms are given below.
\begin{itemize}[leftmargin=1.6 cm,itemsep=-0pt]
\item [(HLS-0)] \Cref{alg:1} with equal allocation (i.e., $\bm p_n = L/m$ with $\widehat{\bm\alpha}$ estimated using step 9).
\item [(HLS-1)] \Cref{alg:1} with $\bm p_n = \bm p^*_n$.
\item [(HLS-2)] \Cref{alg:1} with $\bm p_n = \bm q^*_n$. The regularization parameter $\delta$ in step 6 is chosen as $\delta = 0.01/m$, and the optimization problem is solved using SLSQP with equal-allocation initialization.  
\item [(ERM)] A standard least-squares approach where each training data point consists of a randomly sampled $x\sim\mu$, and a single noisy evaluation $y(x)$ associated with $x$. 
\end{itemize}

For ERM, one may alternatively sample $x$ from a different measure (e.g., $\nu$) and use a weighted $\ell_2$-loss objective in optimization. In such circumstances, the approximation error has a similar dependence on the noise magnitude $\sigma(x)$ as the standard least squares when $L$ is large (i.e., $L\gg n$), which is the regime of interest. Therefore, we do not use further weighting procedures in ERM. 
When comparing the above methods, we fix the total number of evaluations $L$ to be the same to ensure equal comparison. The error metric for comparison is the MSE $\|\widehat{f}-f\|_{L^2_\mu}^2$. 

\subsection{Multivariate function approximation}

Consider the multivariate polynomial approximation problem of the function 
\begin{align*}
f(x) = z^2_1 z_2\exp(z_1+z_2),\quad\quad x = (z_1, z_2)^\top\in \Omega =[-1,1]^2, 
\end{align*}
subject to noisy observations
\begin{align*}
y(x) = f(x) + \sigma(x)\xi, \quad\quad x\in\Omega,
\end{align*}
where $\sigma(x) = 2(1.001-\|x\|_\infty)^2$ and $\xi\sim N(0, 1)$ is a standard normal random variable. We take $V_n$ as the tensor product space of univariate polynomials over $[-1, 1]$ with degrees no more than $D=6$ and the reference measure $\mu$ as the uniform measure on $[-1, 1]^2$, i.e., $V_n = \mathrm{span}\{z^i_1z^j_2: 0\leq i, j\leq D\}$ and $n=\dim(V_n) = (D+1)^2 = 49$. A convenient choice of orthonormal basis in $V_n$ is the tensor product of univariate Legendre polynomials with a similar degree constraint. We use $m=3n$ sample points.

\begin{figure}[htbp]
  \centering 
\begin{subfigure}{0.32\textwidth}{\includegraphics[width=\linewidth, trim={0.6cm 1.3cm 4.5cm 4.2cm},clip]{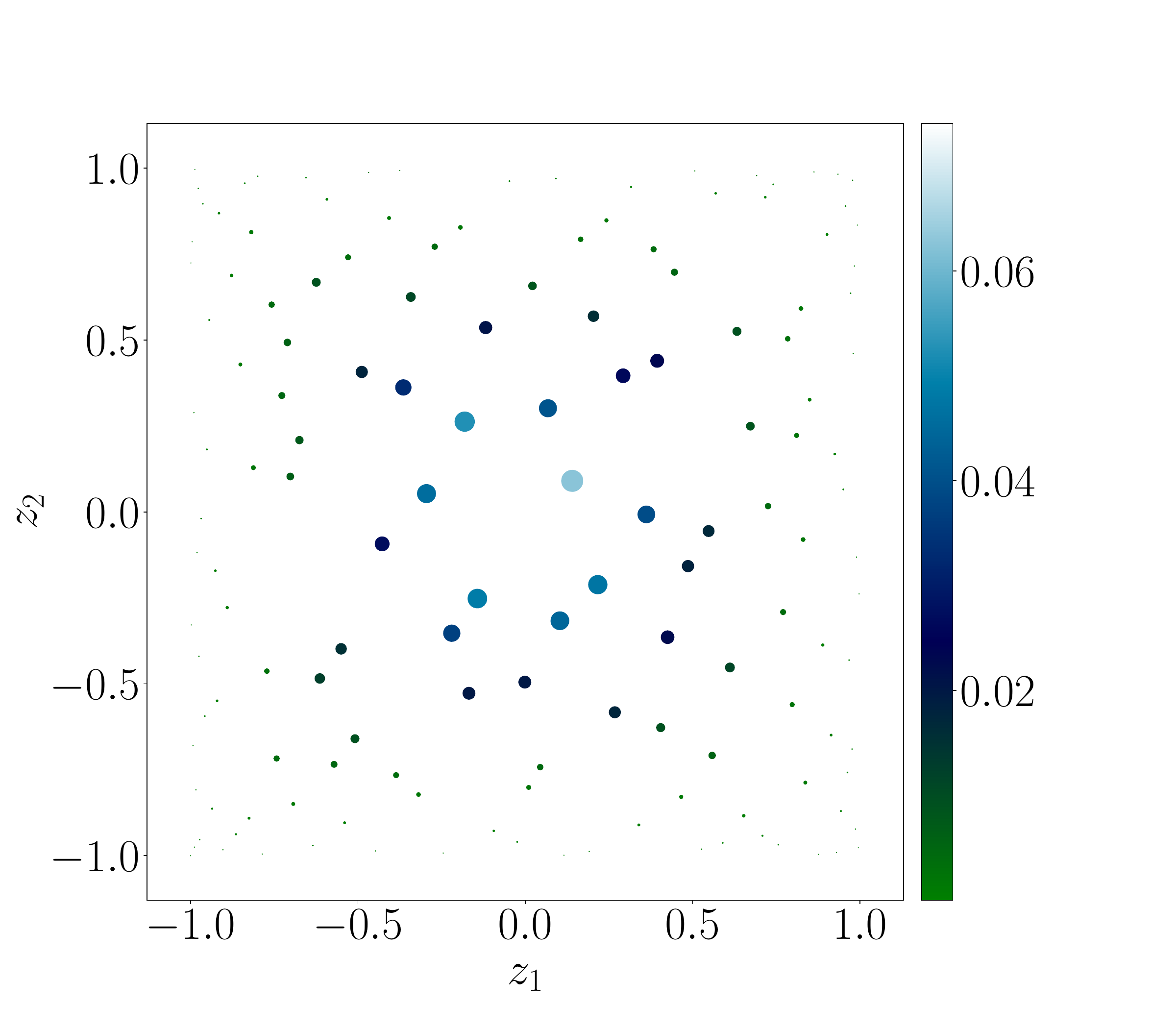}}\caption{}\label{1:(a)}
\end{subfigure} 
\hfill
\begin{subfigure}{0.32\textwidth}{\includegraphics[width=\linewidth, trim={0.6cm 1.3cm 4.5cm 4.2cm},clip]{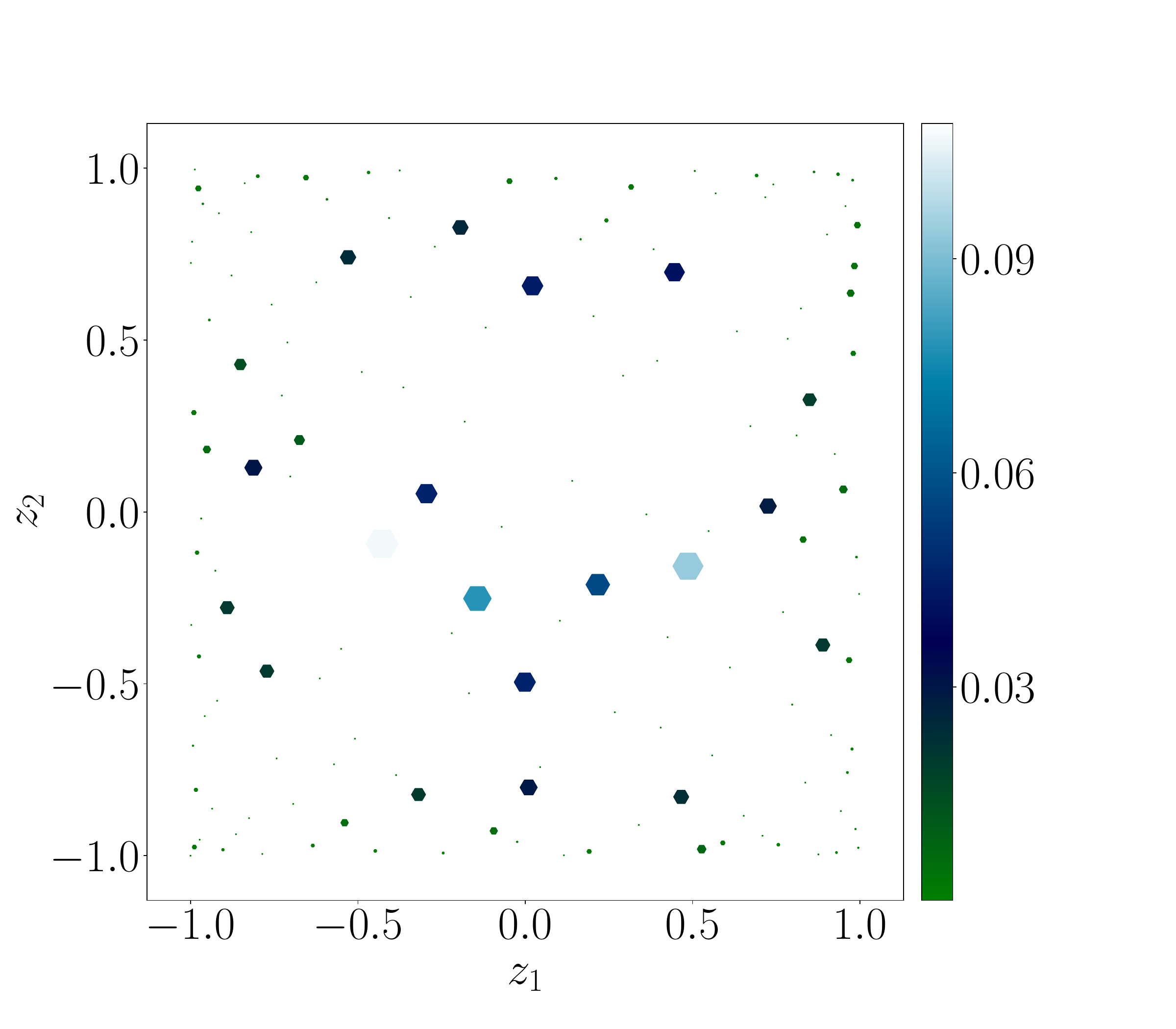}}\caption{}\label{1:(b)}
\end{subfigure} 
\hfill
\begin{subfigure}{0.32\textwidth}{\includegraphics[width=\linewidth, trim={0.0cm 1.2cm 4cm 4.2cm},clip]{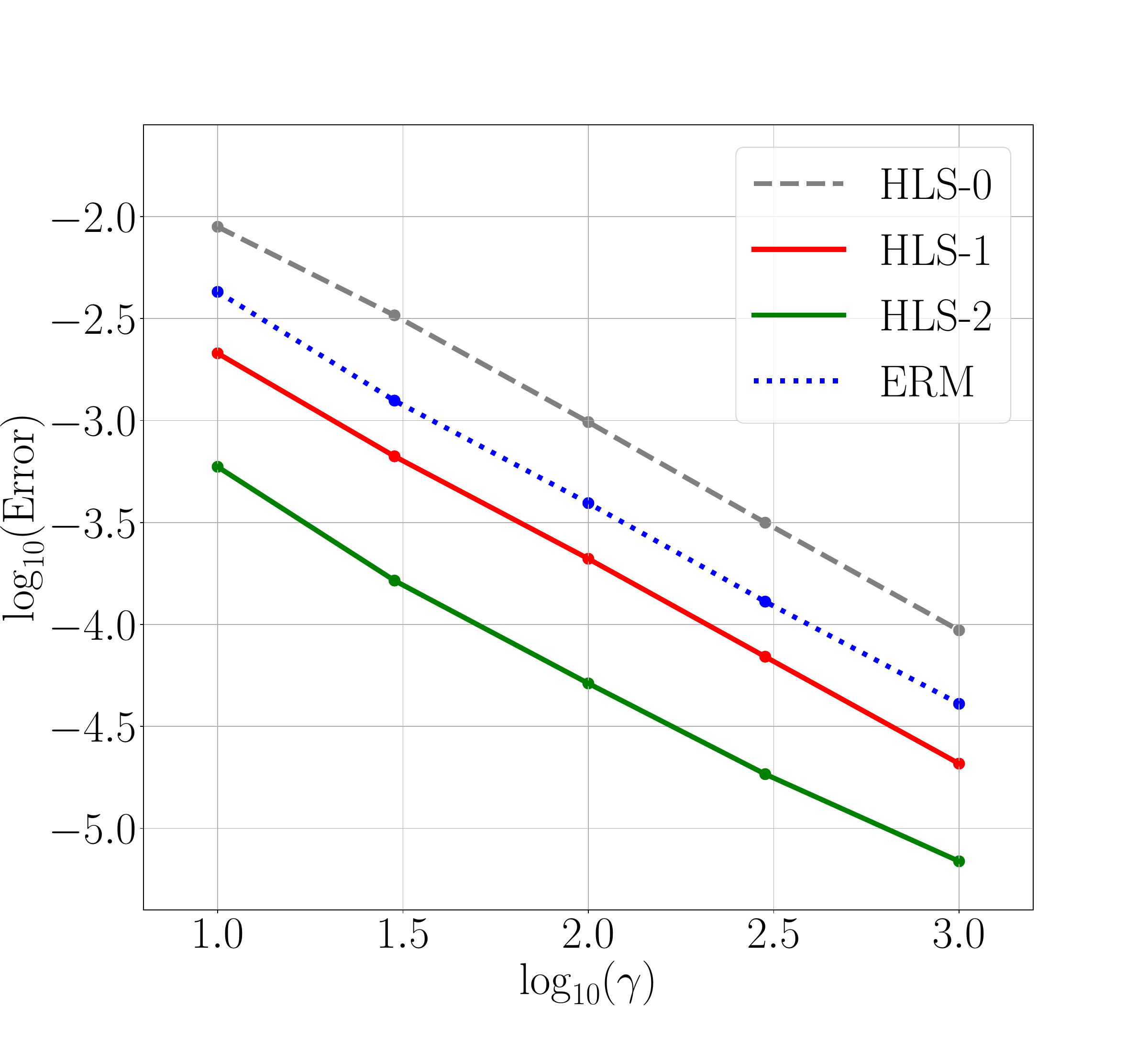}}\caption{}\label{1:(c)}
\end{subfigure} 
\caption{(a)-(b): Scatterplot of the estimated allocation vectors $\bm p_n^*$ (a) and $\bm q_n^{*}$ (b) based on the estimated $\sigma^2(x)$ using $R=50$ MC samples. The allocation weight of each point is indicated by its color and size. (c): Mean squared error of the estimated functions using HLS-0, HLS-1, HLS-2, and ERM under different values of $\gamma$ (for $D = 6$).} \label{fig:11}
\end{figure}

We first draw $m$ sample points using inverse CDF sampling from the Christoffel sampling density, which is a product measure with this choice of $V_n$. To further reduce errors, rather than using independent, uniformly distributed points, we instead take low-discrepancy quasi-random points  (i.e., Halton sequences with bases 2 and 3) over $[-1,1]^2$. We consider four different strategies to estimate the least-squares approximation of $f$ in $V_n$, namely, HLS-0, HLS-1, HLS-2, and ERM. For HLS-1 and HLS-2. For each $x$, we estimate $\sigma(x)$ from $R = 50$ MC simulations offline and use it to compute the corresponding optimal allocation vectors $\bm p_n^*$ and $\bm q_n^{*}$ in Algorithm \ref{alg:1}. To compare the performance of the four methods, we implement a set of different values of $\gamma$:
\begin{align*}
\gamma\in\{10, 30, 100, 300, 1000\},\quad\quad L\in\{2500, 7500, 25000, 75000, 250000\}.
\end{align*}
For each $\gamma$, $100$ experiments are run to compute the MSE. The results are reported in Figure~\ref{fig:11}. 

Figure~\ref{fig:11} shows the numerical results obtained using the compared methods. As anticipated, both $\bm p_n^*$ and $\bm q_n^{*}$ assign more weight to the sample points close to the origin. The corresponding HLS methods, HLS-1 and HLS-2, as shown in Figure~\ref{1:(c)}, exhibit superior performance compared to both HLS-0 and ERM in terms of the approximation error. Notably, HLS-2 outperforms HLS-1 as the smooth $f$ considered in this example is very well approximated by functions in $V_n$.

\begin{figure}[htbp]
  \centering 
\begin{subfigure}{0.32\textwidth}{\includegraphics[width=\linewidth, trim={0.0cm 1.2cm 4cm 4.2cm},clip]{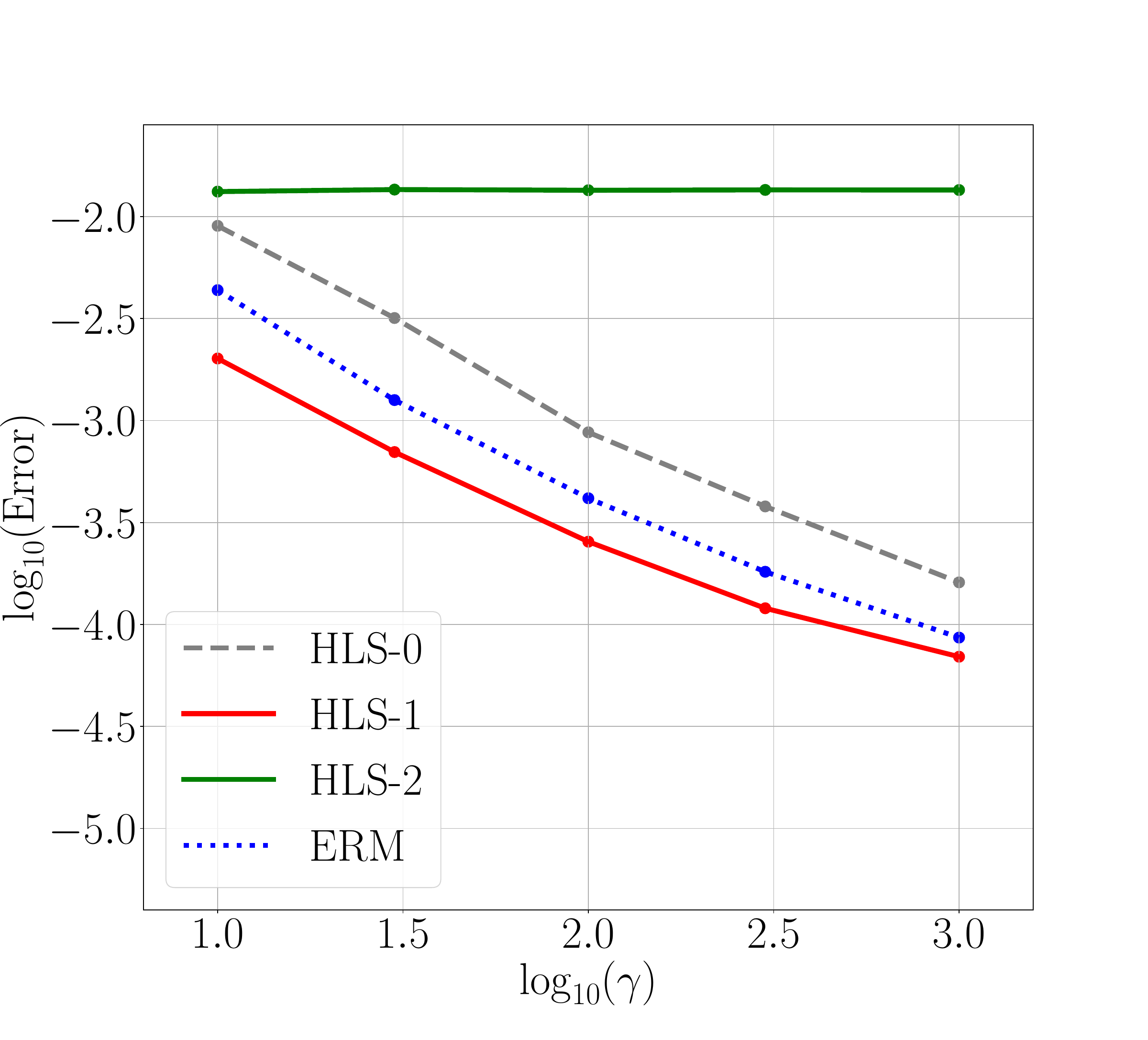}}\caption{}\label{1:(aa1)}
\end{subfigure} 
\hfill
\begin{subfigure}{0.32\textwidth}{\includegraphics[width=\linewidth, trim={0.0cm 1.2cm 4cm 4.2cm},clip]{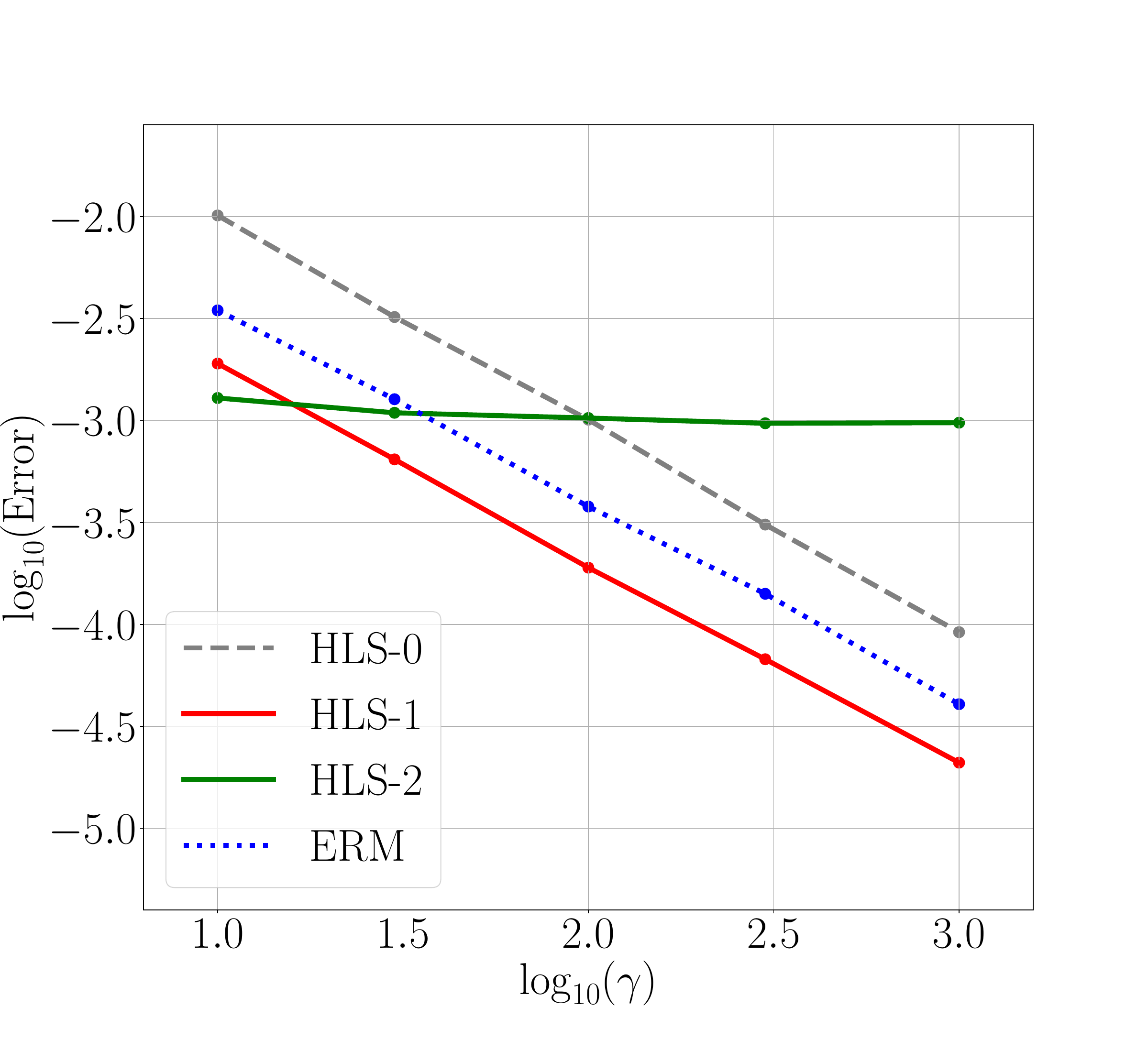}}\caption{}\label{1:(bb1)}
\end{subfigure} 
\hfill
\begin{subfigure}{0.32\textwidth}{\includegraphics[width=\linewidth, trim={0.6cm 1.3cm 4.5cm 4.2cm},clip]{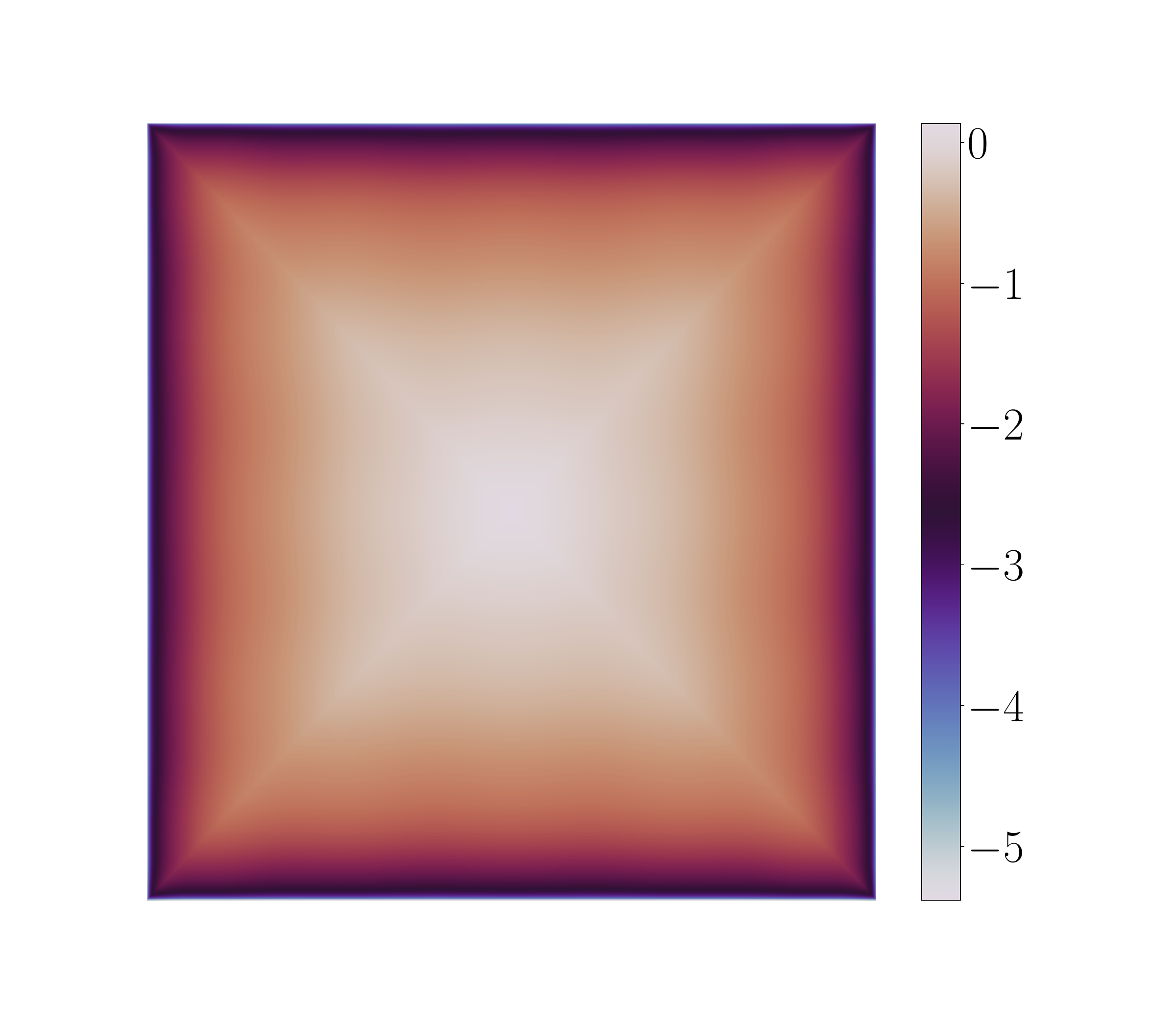}}\caption{}\label{1:(cc1)}
\end{subfigure} 
\caption{(a)-(b): MSE of the estimated functions using HLS-0, HLS-1, HLS-2, and ERM under different values of $\gamma$ for $D = 4$ (a) and $D = 5$ (b). (c): Heatmap of the logarithmic variance factor $\log_{10}(\sigma\sqrt{\Phi_n/n})$ in the case where $D = 6$.} \label{fig:11p}
\end{figure}

We now conduct additional experiments to further investigate the performance of HLS-1, HLS-2, and ERM. First, to examine when HLS-2 outperforms HLS-1, we consider two additional choices of $D$: $D = 4$ and $D = 5$, so that the corresponding $V_n$ have reduced approximation capacity for $f$ as opposed to the previous setup. Keeping $m = 3n = 3(D+1)^2$, we repeat the above simulation and plot the MSE of the estimated functions under different evaluation budgets in Figure~\ref{1:(aa1)}-\ref{1:(bb1)}. For both $D=4$ and $D=5$, the oracle approximation bias $\OPT$ of $V_n$ is relatively large. Since, compared to HLS-1, HLS-2 reduces the estimation variance at the cost of amplifying the approximation bias, its error curve plateaus earlier than HLS-1. The performance of HLS-2 continues to improve as $D$ increases. This suggests that HLS-2 is useful when the approximation bias is much smaller than the estimation variance. 

Second, to understand the accuracy of HLS-1 to ERM, we fix $D = 6$ and plot the function $\sigma\sqrt{\Phi_n/n}$ that appeared in the variance factor in \eqref{p1-1}-\eqref{tired} in \Cref{thm:main} in Figure~\ref{1:(cc1)}. According to \eqref{jgutis}, we also compute the ratio $\|\sigma\sqrt{\Phi_n}\|^2_{L^1_\mu}/\|\sigma\sqrt{\Phi_n}\|^2_{L^2_\mu}\approx 0.453$, which measures the expected efficiency gain of HLS-1 over ERM before the error reaches the order of $\OPT$. In this example, the average ratio of the MSE of HLS-1 and ERM under the five budgets is $0.522$. For the tested budgets in the case $D=6$, the MSE decays linearly, so we expect the variance term to dominate in the MSE. These observations agree qualitatively and quantitatively with the error bounds in \Cref{thm:main}. 

\subsection{Basket options in a bivariate Black--Scholes model}\label{sec:bspr}

Basket options are extensions of single-underlier options, such as European calls or puts, where instead of the single asset underlier, a linear combination of a group of assets is used. In particular, one could use a weighted basket where all the coefficients are positive, or one could use the difference of two assets or two weighted baskets, commonly known as spreads. In this example, we consider spreads under a two-dimensional Black--Scholes setting, where the price vector $S_t = (S^{(1)}_t, S^{(2)}_t)^\top$ follows the following stochastic differential equations:
\begin{equation*}
\begin{aligned}
\d S^{(1)}_t &= rS_t^{(1)} \d t + \sigma^{(1)} S_t^{(1)} \d W^{(1)}_t\\
\d S^{(2)}_t &= rS_t^{(2)} \d t + \sigma^{(2)} S_t^{(2)} \d W^{(2)}_t\\
\E[\d W^{(1)}_t \d W^{(2)}_t] &= \rho \d t
\end{aligned},
\end{equation*}
where $r, \sigma^{(i)}$ are respectively the constant instantaneous rate and volatilities of asset $i$ ($i=1, 2$), and $(W^{(1)}_t, W_t^{(2)})^\top$ is a Brownian motion in $\R^2$ whose increments have constant correlation $\rho$.
A call spread on $S_t^{(1)}$ and $S_t^{(2)}$ with maturity $T$ and strike $K$ has payoff $Y=\max\{S_T^{(1)}-S_T^{(2)}-K, 0\}$, and its price at $t=0$ is 
\begin{align*}
f(T, K, \sigma^{(1)}, \sigma^{(2)}, \rho) \coloneqq e^{-rT}\E[Y\,|\, T, K, \sigma^{(1)}, \sigma^{(2)}, \rho].
\end{align*}
In the following, we fix $r = 0.03$ and $S_0 = (100, 96)^\top$, similar to the setup in \cite{olivares2014note}. 
Our goal is to estimate $f$ as a function of $x = (T, K, \sigma^{(1)}, \sigma^{(2)}, \rho)^\top$ over the target domain $\Omega = [0, 1]\times [0, 50]\times [0, 0.5]\times [0, 0.5]\times [-1, 1]\subset\R^5$. 

This problem belongs to the setting considered in \Cref{sec:rs}, with $g = e^{-r T} Y$. As such, we approximate $f$ using random subspaces $V_n$ generated by random basis functions from $n=100$ MC samples. The choice of $n$ is convenient for balancing computational intensity and approximation accuracy. The random basis functions can be explicitly expressed in this example. Given the standard bivariate normal variables $Z_i: =(z_1(\omega_i), z_2(\omega_i))^\top\sim N(0, \bm I_2)$ where $\omega_i$ denotes the $i$th random seed, the $i$th random basis function in $V_n$ can be expressed using explicit solutions of geometric Brownian motion:
{\begin{align*}
&g(T, K, \sigma^{(1)}, \sigma^{(2)}, \rho; \omega_i) =\max\bigg\{S_0^{(1)}\exp\left(-\frac{(\sigma^{(1)})^2T}{2} + \sigma^{(1)}\sqrt{T}z_1(\omega_i)\right)-\\
&S_0^{(2)}\exp\left(-\frac{(\sigma^{(2)})^2T}{2} + \sigma^{(2)}\sqrt{T}(\rho z_1(\omega_i)+\sqrt{1-\rho^2}z_2(\omega_i))\right)-Ke^{-rT}, 0\bigg\}. 
\end{align*} 
For general stochastic processes, the form of $g$ may not be explicit but could be approximately constructed by standard numerical methods.

In this example, both $g$ and $f$ are positive whereas the least-squares approximant in general is not. To preserve the positivity of the estimation, we take an additional step described in \Cref{sec:convls} where we project the estimated function to the set of nonnegative linear combinations of $\{g(\cdot; \omega_i)\}_{i\in [n]}$, which is a closed convex subset of $V_n$. 

For (approximate) Christoffel sampling, we discretize $\Omega$ using $Q = 2^{16}$= 65,536 quasi-random points generated by a randomly scrambled Sobol' sequence. This results in a matrix of size $Q\times n$, from which a discrete orthonormal basis is obtained through QR decomposition. Using the discrete orthonormal basis, we adaptively select a minimum of $m$ sample points using Christoffel sampling with boosting \cite{haberstich2022boosted} over $50$ experiments to ensure the condition number of the weighted design matrix is less than $2.5$. The value of $m$ is random and slightly fluctuates around 500. For the selected sample points, we reuse the existing samples in $V_n$ to estimate their conditional variances, which are then employed as input to calculate the weight vectors $\bm p^*_n$ and $\bm q^{*}_n$ in Algorithm \ref{alg:1}. As a result, no additional sampling or evaluation is required to compute conditional variances. 

For approximation, we set $L= 5\times 10^5$. We apply HLS and ERM to compute the approximation of $f$ in $V_n$. To evaluate the performance of each method, we uniformly sample $10^3$ points from $\Omega$ and fix and use them as the test dataset. The errors of the estimated functions are computed in the squared $L^2_\mu$ norm, with an oracle value of $f$ computed using MC estimates with $5\times 10^5$ samples. Since $V_n$ is random, we repeat the experiment for $100$ different realizations of $V_n$. The summary statistics are reported in Figure \ref{fig:1}. Moreover, we plot the estimated coefficients vector $\widehat{\bm\alpha}$ given by HSL-1 and HSL-2, each with its coordinates sorted in increasing order in the first experiment in the constrained case. 

\begin{figure}[htbp]
  \centering
\begin{subfigure}{0.32\textwidth}{\includegraphics[width=\linewidth, trim={0.0cm 2.5cm 4cm 4.5cm},clip]{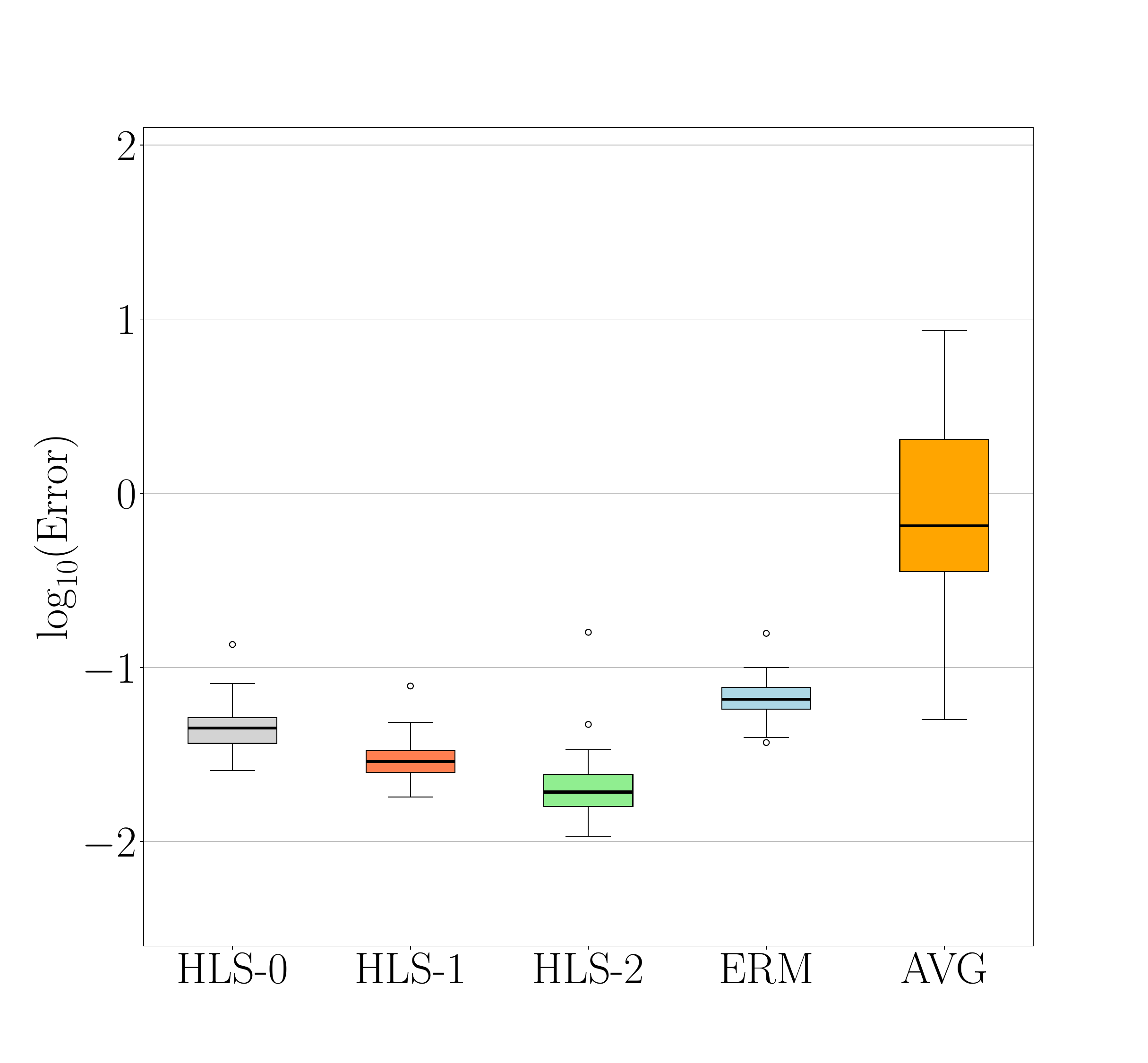}}\caption{}\label{2:(b)}
\end{subfigure}
\hfill 
\begin{subfigure}{0.32\textwidth}{\includegraphics[width=\linewidth, trim={0.0cm 2.5cm 4cm 4.5cm},clip]{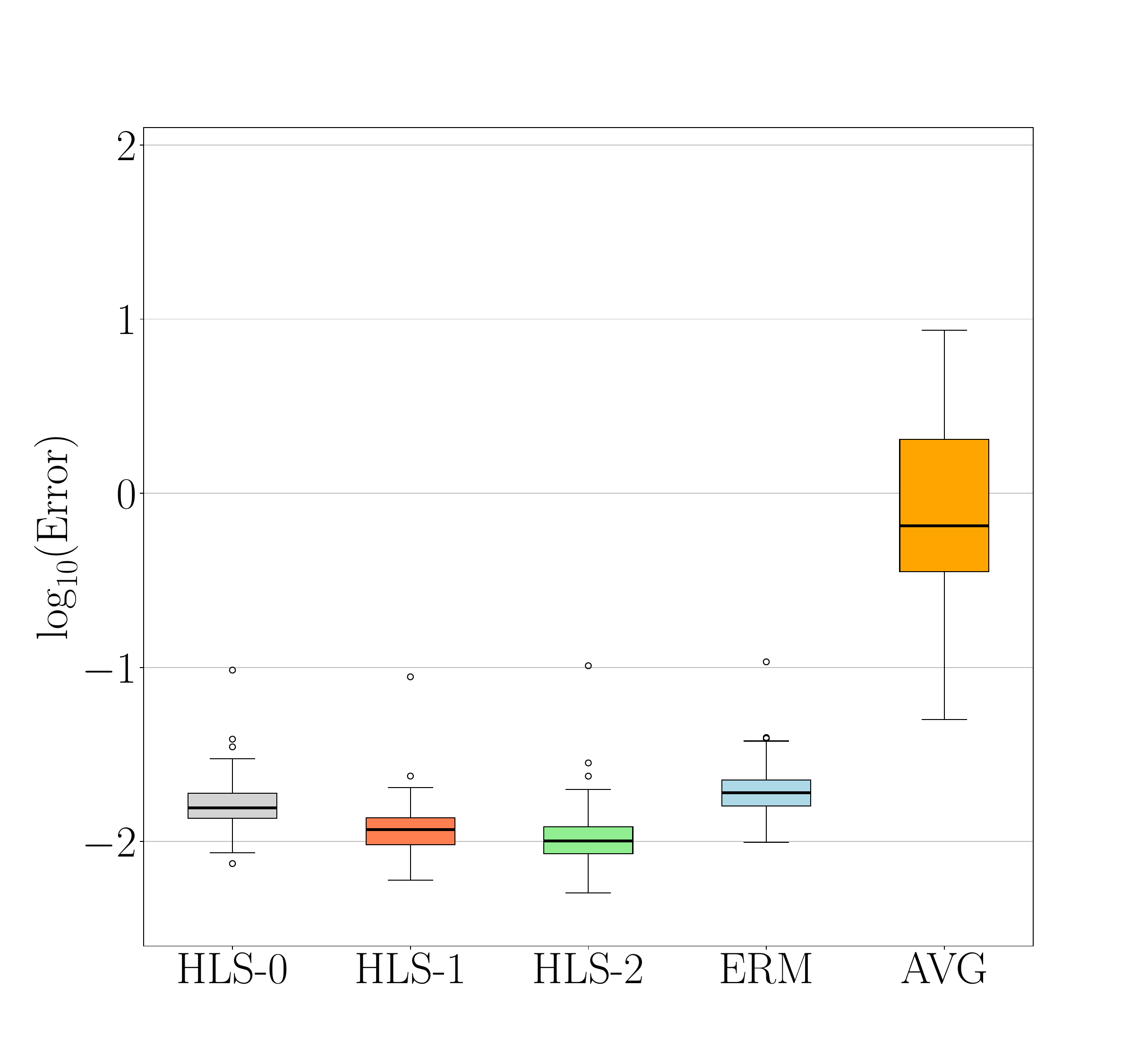}}\caption{}\label{2:(c)}\end{subfigure} 
\hfill 
\begin{subfigure}{0.32\textwidth}{\includegraphics[width=\linewidth, trim={0.0cm 2.5cm 4cm 4.5cm},clip]{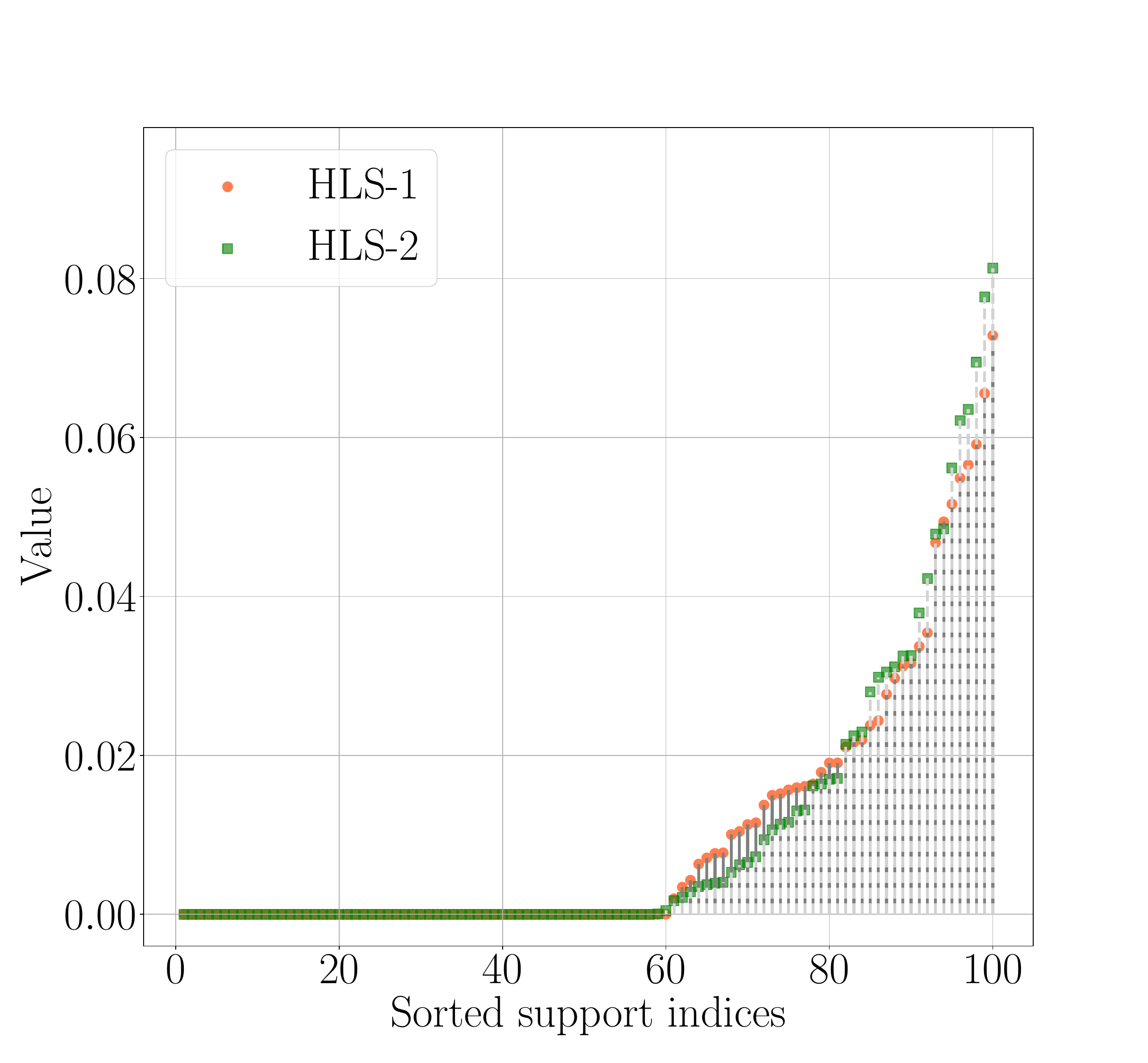}}\caption{}\label{2:(b1)}\end{subfigure} 
\caption{
(a)-(b): Boxplots of the $\log_{10}$(Error) of the estimated least-squares approximant given by HLS-0, HLS-1, HLS-2, ERM, and the average of the random basis functions in $V_n$ (AVG) over the 100 experiments on the test data in the regular setting (a) and the projected setting (b). (c): Sorted estimated coefficient vectors $\widehat{\bm\alpha}$ given by HLS-1 and HLS-2 in the last experiment in the projected setting. }
\label{fig:1}
\end{figure}

Figure \ref{fig:1} contains the simulation results of using HLS-0, HLS-1, HLS-2, and ERM to approximate the spread call function $f(T, K, \sigma^{(1)}, \sigma^{(2)}, \rho)$ in a bivariate Black--Scholes model. After applying projection that preserves the positivity of solutions, all methods demonstrate improved accuracy on the test set compared to the regular least-squares setting. Additionally, both HLS-1 and HSL-2 outperform their uniformly weighted counterpart HLS-0 in both settings, with HSL-2 yielding a slightly more optimal result than HLS-1 on average, as depicted in Figure~\ref{2:(b)}-\ref{2:(c)}. Furthermore, Figure~\ref{2:(b1)} shows that the estimated coefficient vectors $\widehat{\bm\alpha}$ with respect to the random basis demonstrate additional sparsity structure after projection.

In contrast to the example in the previous section, ERM has the worst performance and requires more computational time. Over the 100 experiments, the average time (for both regular and projected) of implementing HLS-1, including the grid discretization, QR decomposition, Christoffel sampling with boosting, and function evaluation, is around 4 seconds. The average time for HLS-2, which requires solving an additional constrained optimization problem, is around 7 seconds, with an average of 15 SLSQP iterations. The average time for ERM, involving the construction of a large design matrix and solving the corresponding least-squares problem, is around 24 seconds. This highlights the advantage of structured design and computation for both accuracy and efficiency when employing least squares for noisy function approximation.

By comparing the approximation results of HLS/ERM and AVG, we observe that although the average of the random basis functions in $V_n$ provides a poor estimate for $f$, with high probability, there exists an element within $V_n$ that can sufficiently approximate $f$. This finding further supports the results in \Cref{thm:984}.

\section{Conclusion}

We developed a hybrid least-squares method for noisy function approximation, with a special focus on the scenario where noise is large. Such situations are commonplace in stochastic simulations such as computational finance. The proposed algorithm combines Christoffel sampling with an additional step that allocates the function evaluation budget based on certain experimental design criteria, utilizing conditional variance information. We showed that the proposed algorithms enjoy both improved accuracy and efficiency compared to least-squares approaches that do not leverage conditional variance information. We also demonstrated that the proposed algorithms can be applied to the constrained setting with minor modifications. Furthermore, for applications where the noise across the domain depends on a set of shared random variables, we proposed a sequence of adaptive random subspaces to approximate the target function and analyzed its approximation capability. Through several numerical experiments, we showed that the proposed hybrid method demonstrates both effectiveness and efficiency in handling noisy function approximation problems. In particular, the reweighted approach HLS-2 may exhibit superiority when the approximation subspace $V_n$ is known a priori to be sufficiently expressive, whereas the non-reweighted approach HLS-1 is a safe choice that provably improves upon standard randomized least-squares methods. 

Although the proposed methods appear promising based on initial simulation studies, the choice of the regularization parameter $\delta$ in the reweighted allocation is not yet fully understood, which may limit its practical applicability. One possible strategy is to select $\delta$ just large enough so that $\mathcal P_m(\delta)$ contains the non-reweighted optimal allocation as a feasible solution. Moreover, in certain applications, derivative information may be available in addition to function evaluations. We leave the investigation of these directions for future work.

\bibliographystyle{siamplain}
\bibliography{ref}


\begin{appendix}
\section{Optimal allocation}

\subsection{Proof of \Cref{lm:a1}}\label{appd:1}
By a direct computation, 
\begin{align}
\E_{\bm y}[\|\widehat{\bm\alpha}(\bm p)-\bar{\bm\alpha}\|_2^2\,|\, \XX]=&\ \E_{\bm y}[\|(\bm V^\top\bm W\bm V)^{-1}\bm V^\top\bm W^{\frac{1}{2}}(\bm y-\bm f)\|_2^2\,|\, \XX]\label{eq1}\\
\leq&\  \|(\bm V^\top\bm W\bm V)^{-1}\|_2^2\cdot\E_{\bm y}[\|\bm V^\top\bm W^{\frac{1}{2}}(\bm y-\bm f)\|_2^2\,|\, \XX]\nonumber\\
\stackrel{\eqref{n:design}}{=}&\  \|(\bm V^\top\bm W\bm V)^{-1}\|_2^2\cdot\tr(\bm V^\top\bm W^{\frac{1}{2}}\bm\Sigma(\bm p)\bm W^{\frac{1}{2}}\bm V)\nonumber\\
=&\ \|(\bm V^\top\bm W\bm V)^{-1}\|_2^2\cdot G(\bm p)\nonumber, 
\end{align}
where 
\begin{align}
G(\bm p) \coloneqq \frac{1}{L}\sum_{i\in [m]}\frac{w^2(x_i)\sigma^2(x_i)\Phi_n(x_i)}{m^2p_i}.\label{myG}
\end{align}
By a similar argument, 
\begin{align}
\E_{\bm y}[\|\widehat{\bm\alpha}(\bm p)-\bar{\bm\alpha}\|_2^2\,|\, \XX]\geq \frac{G(\bm p)}{\|\bm V^\top\bm W\bm V\|_2^2}.\label{eq2}
\end{align}
Therefore, $\E_{\bm y}[\|\widehat{\bm\alpha}(\bm p)-\bar{\bm\alpha}\|_2^2\,|\, \XX]$ and $G(\bm p)$ are equivalent up to a factor $\cond(\bm V^\top\bm W\bm V)^2$. Since $G(\bm p)$ is a strictly convex function of $\bm p$ in $\mathcal P_m$ that diverges on the boundary, a unique minimizer $\bm p_n^*$ exists and is given by \eqref{neyman}, with optimal value
\begin{align}
G(\bm p_n^*) = \frac{1}{L}\left(\frac{1}{m}\sum_{i\in [m]}w(x_i)\sigma(x_i)\sqrt{\Phi_n(x_i)}\right)^2. \label{mygstar}
\end{align}
Denoting $\bm p_n$ an optimal solution to \eqref{lsjs}, 
\begin{align*}
\E_{\bm y}[\|\widehat{\bm\alpha}(\bm p_n^*)-\bar{\bm\alpha}\|_2^2\,|\, \XX]&\stackrel{\eqref{eq1}}{\leq}\|(\bm V^\top\bm W\bm V)^{-1}\|_2^2\cdot G(\bm p_n^*)\leq \|(\bm V^\top\bm W\bm V)^{-1}\|_2^2\cdot G(\bm p_n)\\
&\stackrel{\eqref{eq2}}{\leq}\cond(\bm V^\top\bm W\bm V)^2\cdot \E_{\bm y}[\|\widehat{\bm\alpha}(\bm p_n)-\bar{\bm\alpha}\|_2^2\,|\, \XX].
\end{align*}

\subsection{Proof of \Cref{mkjh}}\label{app:conxh}
To show \eqref{H-conv+} is a convex optimization problem, note that the feasible set $\mathcal P_m(\delta)$ is convex, so it remains to verify the convexity of the objective $H(\bm p)$. To do this, we first verify that the objective $H(\bm p)$ is well-defined. 
Recall that $H(\bm p) = \tr((\bm U(\bm p))^{-1})$ in \eqref{H-conv}. Under the column-rank assumption on $\bm W^{\frac{1}{2}}\bm V$, $\bm U(\bm p)$ is invertible for $\bm p\in\mathcal P_m(\delta)$. Therefore, $H(\bm p)<\infty$. To establish convexity, we take $\bm p, \bm p'\in \mathcal P_m(\delta)$ and $\lambda\in [0,1]$. In this case, note that $\bm U(\bm p)$ is linear in $\bm p$ and the function $x\mapsto 1/x$ is operator convex \cite{bendat1955monotone}, that is, for two feasible $\bm p, \bm p'$ and $\lambda\in [0,1]$, we have
\begin{align*}
\left(\bm U(\lambda\bm p +(1-\lambda)\bm p')\right)^{-1}=\left(\lambda \bm U(\bm p) + (1-\lambda) \bm U(\bm p')\right)^{-1}\preceq\lambda \bm U(\bm p)^{-1} + (1-\lambda) \bm U(\bm p')^{-1}.
\end{align*}
Taking the trace on both sides yields the desired convexity. Consequently, an optimal solution exists with a finite objective value. 

We now show that there exists an optimal solution $\bm q^{*}_n$ with at most $(n^2+n)/2$ components greater than $\delta$. Let $R^\top_i$ denote the $i$th row vector of $\bm W^{\frac{1}{2}}\bm V$. If $|\supp_{\delta}(\bm q_n^{*})|>n(n+1)/2$, then $R_iR^\top_i, i\in\text{supp}_{\delta}(\bm q^{*}_n)$ are linearly dependent. Therefore, there exists a direction $\bm a = (a_1, \ldots, a_m)^\top$ supported on $\text{supp}_{\delta}(\bm q^{*}_n)$ such that  $\sum_{i\in [m]}\frac{a_i}{\sigma^2(x_i)}R_iR^\top_i = 0$. Applying a perturbation argument to $\bm q^{*}_n$ along $\bm a$ yields another optimal solution that has a smaller $\delta$-support. Proceeding with such operations until $|\supp_{\delta}(\bm q_n^{*})|\leq n(n+1)/2$ finishes the proof.

\subsection{Proof of \Cref{lm:a2}}\label{app:conxhl}
With $\widehat{\bm\alpha}$ chosen as the $\bm\Gamma$-reweighted least-squares in \eqref{sdr},  
\begin{align*}
\widehat{\bm\alpha}(\bm p) = (\bm\Gamma\bm W^{\frac{1}{2}}\bm V)^\dagger\bm\Gamma(\bm z_1 + \bm z_2 + \bm y-\bm f)= \bar{\bm\alpha} + (\bm\Gamma\bm W^{\frac{1}{2}}\bm V)^\dagger\bm\Gamma(\bm z_2 + \bm y-\bm f).
\end{align*}
Thus, the $\XX$-conditional MSE can be computed using the bias-variance decomposition: 
\begin{align}
\E_{\bm y}[\|\widehat{\bm\alpha}(\bm p)-\bar{\bm\alpha}\|_2^2\,|\, \XX] &= \|(\bm\Gamma\bm W^{\frac{1}{2}}\bm V)^\dagger\bm\Gamma\bm z_2\|_2^2 + \E\left[\|(\bm\Gamma\bm W^{\frac{1}{2}}\bm V)^\dagger\bm\Gamma(\bm y-\E_{\bm y}[\bm y])\|_2^2\,|\, \XX\right]\nonumber\\
& = \|(\bm\Gamma\bm W^{\frac{1}{2}}\bm V)^\dagger\bm\Gamma\bm z_2\|_2^2 + H(\bm p),\label{mse-tr}
\end{align}
where the last step follows by taking $\bm\Gamma = \bm\Sigma(\bm p)^{-\frac{1}{2}}$.
The first term in \eqref{mse-tr} can be bounded using $\cond(\bm\Sigma(\bm p))$. 
Since $\bm\Gamma\bm W^{\frac{1}{2}}\bm V(\bm\Gamma\bm W^{\frac{1}{2}}\bm V)^\dagger$ is an orthogonal projection,  
\begin{align*}
\|\bm\Gamma\|^2_2\cdot\|\bm z_2\|_2^2\geq\|\bm\Gamma\bm z_2\|_2^2\geq\|\bm\Gamma\bm W^{\frac{1}{2}}\bm V(\bm\Gamma\bm W^{\frac{1}{2}}\bm V)^\dagger\bm\Gamma\bm z_2\|^2_2\geq \frac{\|(\bm\Gamma\bm W^{\frac{1}{2}}\bm V)^\dagger\bm\Gamma\bm z_2\|^2_2}{\|(\bm V^\top\bm W\bm V)^{-1}\|_2\cdot\|\bm\Gamma^{-1}\|^2_2},
\end{align*}
which can be simplified to 
\begin{align}
\|(\bm\Gamma\bm W^{\frac{1}{2}}\bm V)^\dagger\bm\Gamma\bm z_2\|^2_2&\leq\cond(\bm\Gamma^2) \|(\bm V^\top\bm W\bm V)^{-1}\|_2 \|\bm z_2\|_2^2\nonumber\\
& = \cond(\bm\Sigma(\bm p)) \|(\bm V^\top\bm W\bm V)^{-1}\|_2 \|\bm z_2\|_2^2.\label{hr1}  
\end{align}
A direct computation shows,
\begin{align}
\sup_{\bm p\in\mathcal P_{m}(\delta)}\cond(\bm\Sigma(\bm p)) \stackrel{\eqref{n:design}}{=}\sup_{\bm p\in\mathcal P_{m}(\delta)}\frac{\max_{i\in [m]}\frac{w(x_i)\sigma^2(x_i)}{mp_i}}{\min_{i\in [m]}\frac{w(x_i)\sigma^2(x_i)}{mp_i}}\leq\frac{J_n}{\delta},\label{conG}
\end{align}
where $J_n$ is defined in \Cref{lm:a2}. 
Substituting \eqref{hr1} and \eqref{conG} into \eqref{mse-tr} yields  
\begin{align}
H(\bm p)\leq\E_{\bm y}[\|\widehat{\bm\alpha}(\bm p)-\bar{\bm\alpha}\|_2^2\,|\, \XX]\leq \frac{J_n}{\delta}\|(\bm V^\top\bm W\bm V)^{-1}\|_2\|\bm z_2\|_2^2+H(\bm p),\quad\quad\forall\bm p\in\mathcal P_m(\delta).\label{ail}
\end{align}
Note that the first term of the upper bound in \eqref{ail} is independent of $\bm p$. As a result, for any $\bm p_n\in\argmin_{\bm p\in\mathcal P_m(\delta)}\E_{\bm y}[\|\widehat{\bm\alpha}(\bm p)-\bar{\bm\alpha}\|_2^2\,|\, \XX]$, 
\begin{align*}
\E_{\bm y}[\|\widehat{\bm\alpha}(\bm q_n^{*})-\bar{\bm\alpha}\|_2^2\,|\, \XX]&\leq\frac{J_n}{\delta}\|(\bm V^\top\bm W\bm V)^{-1}\|_2\|\bm z_2\|_2^2+H(\bm q_n^{*})\\
&\leq \frac{J_n}{\delta}\|(\bm V^\top\bm W\bm V)^{-1}\|_2\|\bm z_2\|_2^2+H(\bm p_n)\\
&\stackrel{\eqref{mse-tr}}{\leq}\frac{J_n}{\delta}\|(\bm V^\top\bm W\bm V)^{-1}\|_2\|\bm z_2\|_2^2+\E_{\bm y}[\|\widehat{\bm\alpha}(\bm p_n)-\bar{\bm\alpha}\|_2^2\,|\, \XX].
\end{align*}

\section{Hybrid-least squares algorithms and error bounds}
\subsection{Proof of \Cref{thm:main}}\label{app:main}
Assume $n\geq 2$. We first prove the case where $\bm p_n = \bm p_n^*$. 
Consider the intermediate least squares problem
\begin{align}
&\bm W^{\frac{1}{2}}\bm V\bm\alpha = \E_{\bm y}[\bm y\,|\,\XX].\label{1}
\end{align}
According to \cite[Theorem 2]{cohen2017optimal}, if $m\gtrsim K_n(w)\log K_n(w)$ where $K_n(w) \coloneqq\|\Phi_n(x)w(x)\|_{L^\infty_\mu}\geq n$, then with probability at least $1-n^{-2}$, $\bm W^{\frac{1}{2}}\bm V$, viewed as a mapping from $V_n$ to $\R^m$: $\sum_{i\in [n]}\alpha_iv_i\mapsto \bm W^{\frac{1}{2}}\bm V\bm\alpha$, is an $(1\pm 0.1)$-subspace embedding: 
\begin{align}
\Lambda(\bm W^{\frac{1}{2}}\bm V)\subseteq[0.9, 1.1]\Rightarrow\cond(\bm V^\top\bm W\bm V)\leq \left(\frac{1.1}{0.9}\right)^2<\frac{3}{2}. \label{2}
\end{align}
The lower bound $K_n(w) = n$ is attained when $w(x)$ is the induced measure in \Cref{alg:1}.  

We now denote $\AA$ the probabilistic event in \eqref{2}, i.e., 
\begin{align}
\AA\coloneqq\left\{\XX\in\Omega^{m}: \Lambda(\bm W^{\frac{1}{2}}\bm V)\subseteq[0.9, 1.1]\right\}, \quad\quad\P(\AA)>1-n^{-2}. \label{mist}
\end{align}
Let $\bm\alpha^*$ be the solution to \eqref{alphaoracle} and $\bar{\bm\alpha}$ be the solution to \eqref{1}.
Continuing to follow the proof of \cite[Theorem 2]{cohen2017optimal}, we obtain,
\begin{align}
\E_{\XX}\left[\left\|f^*-\bar{f}\right\|^2_{L^2_\mu}\,|\,\AA\right]\leq \frac{cn}{2m}\cdot\OPT\label{ineq:1},
\end{align}
where $f^* = \sum_{i\in [n]}\alpha_i^*v_i$, $\bar{f} = \sum_{i\in [n]}\bar{\alpha}_iv_i$, and $c>0$ is some absolute constant. 

We now appeal to the results in \Cref{sec:3.1}. 
Taking $\bm p=\bm p^*_n$ in \eqref{eq1} yields that
\begin{align}
\E_{\XX, \bm y}[\|\widehat{\bm\alpha}-\bar{\bm\alpha}\|_2^2\,|\,\AA]&\stackrel{\eqref{eq1}}{\leq}\E_{\XX}[\|(\bm V^\top\bm W\bm V)^{-1}\|_2^2\cdot G(\bm p^*_n)\,|\,\AA]\stackrel{\eqref{2}}{\leq}\frac{3}{2}\E_{\XX}[G(\bm p^*_n)\,|\,\AA]\nonumber\\
&= \frac{3}{2}\frac{\E_{\XX}[G(\bm p^*_n)\mathbb I_{\AA}]}{\P(\AA)}\stackrel{\eqref{mist}}{\leq} 2\E_{\XX}[G(\bm p^*_n)].\label{tpo}
\end{align}
The expectation $\E_{\XX}[G(\bm p^*_n)]$ can be explicitly computed using \eqref{mygstar}:
\begin{align}
\E_{\XX}[G(\bm p^*_n)] & = \E_{\XX}\left[\frac{1}{L}\left(\frac{1}{m}\sum_{i\in [m]}w(x_i)\sigma(x_i)\sqrt{\Phi_n(x_i)}\right)^2\right]\nonumber\\
& =  \frac{1}{L}\left[\frac{1}{m}\E_{\XX}[w^2(x_1)\sigma^2(x_1)\Phi_n(x_1)] + \left(1-\frac{1}{m}\right)\E_{\XX}\left[w(x_1)\sigma(x_1)\sqrt{\Phi_n(x_1)}\right]^2\right]\nonumber\\
& \stackrel{\eqref{chris}}{=} \frac{n}{L}\left[\frac{1}{m}\|\sigma\|^2_{L^2_\mu} + \left(1-\frac{1}{m}\right)\left\|\sigma\sqrt{\frac{\Phi_n}{n}}\right\|^2_{L^1_\mu}\right]\label{lo904}.
\end{align}

Combining \eqref{ineq:1}, \eqref{lo904} and applying the Pythagorean theorem and Cauchy--Schwarz inequality, we have
\begin{align}
\E_{\XX, \bm y}\left[\left\|\widehat{f}-f\right\|^2_{L^2_\mu}\,|\, \AA\right]& = \left\|f-f^*\right\|^2_{L^2_\mu} + \E_{\XX, \bm y}\left[\left\|\widehat{f}-f^*\right\|^2_{L^2_\mu}\,|\, \AA\right]\label{jtr}\\
&\leq \OPT + 2\left(\E_\XX\left[\left\|\bar{f}-f^*\right\|^2_{L^2_\mu}\,|\,\AA\right] + \E_{\XX, \bm y}\left[\left\|\widehat{f}-\bar{f}\right\|^2_{L^2_\mu}\,|\, \AA\right]\right)\nonumber\\
&\stackrel{\eqref{ineq:1}, \eqref{tpo}}{\leq} \left(1+\frac{cn}{m}\right)\OPT + 4\E_{\XX}[G(\bm p^*_n)]\nonumber.
\end{align}
Substituting $\E_{\XX}[G(\bm p^*_n)]$ using \eqref{lo904} yields the desired result. 

The proof for the case $\bm p_n = \bm q_n^{*}$ is similar and we only point out the differences. 
Proceeding with the same event $\AA$ as in the previous case and applying \eqref{ail}, we have 
\begin{align}
\E_{\XX, \bm y}\left[\left\|\widehat{f}-f\right\|^2_{L^2_\mu}\,|\, \AA\right]\stackrel{\eqref{ail}, \eqref{2}}{\leq}\left(1+\frac{cn}{m}\right)\OPT + \frac{3J_n}{2\delta}\E_\XX[\|\bm z_2\|^2_2\,|\,\AA]+4\E_\XX[H(\bm q_n^{*})],
\end{align}
where $\bm z_2$ is defined in \eqref{myz12}. 
The proof is finished by bounding $\E_\XX[\|\bm z_2\|^2_2\,|\,\AA]$ as follows:
\begin{align*}
&\frac{3}{2}\E_\XX[\|\bm z_2\|^2_2\,|\,\AA] = \frac{3}{2}\E_\XX\left[\left\|\E[\bm y]-\bm W^{\frac{1}{2}}\bm V\bar{\bm\alpha}\right\|_2^2\,|\,\AA\right]\leq \frac{3}{2}\E_\XX\left[\left\|\E[\bm y]-\bm W^{\frac{1}{2}}\bm V\bm\alpha^*\right\|_2^2\,|\,\AA\right]\\
&\stackrel{\eqref{mist}}{\leq} 2\E_\XX\left[\left\|\E[\bm y]-\bm W^{\frac{1}{2}}\bm V\bm\alpha^*\right\|_2^2\right]= 2\E_\XX\left[\frac{1}{m}\sum_{i\in [m]}w(x_i)\left(f(x_i)-f^*(x_i)\right)^2\right]= 2\OPT. 
\end{align*}

\subsection{Proof of \Cref{thm:conv}}\label{app:conv}
The proof is similar to \Cref{thm:main} and we only highlight the differences. We first note 
\begin{align*}
\|f-\widehat{f}_c\|^2_{L^2_\mu}\leq 2(\|f-f^*_c\|^2_{L^2_\mu} + \|f^*_c-\widehat{f}_c\|^2_{L^2_\mu}) = 2\OPT_c + 2\|f^*_c-\widehat{f}_c\|^2_{L^2_\mu}.
\end{align*}
To bound $\|f^*_c-\widehat{f}_c\|^2_{L^2_\mu}$, it follows from \Cref{cont} that 
\begin{align*}
\|f^*_c-\widehat{f}_c\|^2_{L^2_\mu} &\stackrel{\eqref{obs}}{=} \|\Pi_{\mathcal C}(f^*)-\Pi_{\mathcal C}(\widehat{f})\|^2_{L^2_\mu}\stackrel{(\text{Lemma}~\ref{cont})}{\leq} \|f^*-\widehat{f}\|^2_{L^2_\mu}\leq 2\left(\|\bar{f}-f^*\|^2_{L^2_\mu} + \|\bar{f}-\widehat{f}\|^2_{L^2_\mu}\right),
\end{align*}
where $\bar{f}$ is the same as in the proof of \Cref{thm:main}. Noting $\OPT\leq\OPT_c$, the rest of the proof is similar to the proof of \Cref{thm:main}.

\subsection{Analysis of pilot variance estimation}\label{app:cost}
The analysis of both \Cref{thm:main} and \Cref{thm:conv} assumes that $\sigma^2(x)$ is given. If not, an additional evaluation budget is required to estimate $\sigma^2(x)$ on $\XX$, which incurs extra cost and leads to a loss of efficiency. In this section, we show that as long as the estimated variance (based on sample variance estimators with $R$ samples), denoted by $\widehat{\sigma}^2(x)$, is accurate up to a small constant factor of the true variance on $\XX$, the corresponding optimal allocation vectors (namely, $\widehat{\bm p}^*_n$/$\widehat{\bm q}^*_n$ for the non/reweighted cases) are near-optimal. Moreover, under suitable moment assumptions, the required variance estimation accuracy can be achieved at a cost negligible compared to the total evaluation budget. 
\begin{theorem}\label{thm:pert}
Assuming that there exists $\kappa<1$ such that 
\begin{align}
\left |\frac{\widehat{\sigma}^2(x_i)}{\sigma^2(x_i)}-1\right|\leq \kappa, \quad\quad i\in [m], \label{bushdge}
\end{align}
then 
\begin{align}
G(\widehat{\bm p}^*_n)\leq\frac{1+\kappa}{1-\kappa}G(\bm p_n^*), \quad\quad H(\widehat{\bm q}^*_n)\leq\frac{1+\kappa}{1-\kappa}H(\bm q_n^*), \label{subbddo}
\end{align}
where $G$ (c.f.~\eqref{myG}) and $H$ (c.f.~\eqref{H-conv}) are the allocation objectives optimized in the non-reweighted and reweighted settings, respectively. 
 
Furthermore, if the normalized noise sequence $\{\e(x)/\sigma(x)\}_{x\in\Omega}$ is uniformly subgaussian, then for $R\gtrsim \log m/\kappa^2$ (where the implicit constant depends on the uniform subgaussian norm of the noise sequence), the conditions in \eqref{bushdge} hold with probability at least $1-m^{-2}$. 
\end{theorem}

\eqref{subbddo} characterizes the suboptimality of the estimated allocation vectors relative to the optimal ones. To achieve this bound with high probability, one can choose $R \gtrsim \log m / \kappa^2$, which is effectively of order $1/\kappa^2$. In practice, we take $L = \gamma m$, so the cost of this pilot variance estimation becomes negligible if $\gamma\gg 1/\kappa^2$. When the subgaussian assumption is violated, typically due to points with near-vanishing variances, the empirical variance estimates may be zero. While we do not have a rigorous theoretical analysis in this case, an effective empirical remedy is to add a small positive constant to ensure nonsingularity in the computations.

\begin{proof}
We make the dependence on $\sigma^2$ explicit in $G$ and $H$. For the non-reweighted case, since the objective $G(\bm p; \sigma^2)$ has a linear dependence on the sequeunce $\{\sigma^2(x_i)\}_{i\in [m]}$, 
\begin{align*}
\left(1-\kappa\right)G(\bm p; \sigma^2)\leq G(\bm p; \widehat{\sigma}^2)\leq \left(1+\kappa\right)G(\bm p; \sigma^2).
\end{align*}
Consequently, the optimal allocation vector $\widehat{\bm p}^*_n$ with respect to $G(\bm p; \widehat{\sigma}^2)$ satisfies 
\begin{align*}
G(\widehat{\bm p}^*_n; \sigma^2)\leq \frac{1}{1-\kappa}G(\widehat{\bm p}^*_n; \widehat{\sigma}^2)\leq \frac{1}{1-\kappa}G(\bm p_n^*; \widehat{\sigma}^2)\leq\frac{1+\kappa}{1-\kappa}G(\bm p_n^*; \sigma^2). 
\end{align*}
For the reweighted case, recall the notation in \Cref{app:conxh}: 
\begin{align*}
H(\bm p; \sigma^2) = \tr(\bm U(\bm p; \sigma^2)), \quad\quad \bm U(\bm p; \sigma^2) = \sum_{i\in [m]}\frac{p_i}{\sigma^2(x_i)}R_iR_i^\top, 
\end{align*}
where $R^\top_i$ denotes the $i$th row vector of $\bm W^{\frac{1}{2}}\bm V$. Therefore, 
\begin{align*}
\frac{1}{1+\kappa}\bm U\left(\bm p; \sigma^2\right)=\bm U\left(\bm p; \left(1+\kappa\right)\sigma^2\right)\preceq \bm U(\bm p; \widehat{\sigma}^2)\preceq \bm U\left(\bm p; \left(1-\kappa\right)\sigma^2\right) = \frac{1}{1-\kappa}\bm U\left(\bm p; \sigma^2\right).
\end{align*}
Since $x\mapsto 1/x$ is operator monotone decreasing, 
\begin{align*}
\left(1-\kappa\right)\bm U^{-1}\left(\bm p; \sigma^2\right)\preceq \bm U^{-1}(\bm p; \widehat{\sigma}^2)\preceq \left(1+\kappa\right)\bm U^{-1}\left(\bm p; \sigma^2\right).
\end{align*}
Taking trace on both sides yields $\left(1-\kappa\right)H(\bm p; \sigma^2)\leq H(\bm p; \widehat{\sigma}^2)\leq \left(1+\kappa\right)H(\bm p; \sigma^2)$, which implies that
\begin{align*}
H(\widehat{\bm q}^*_n; \sigma^2)\leq \frac{1+\kappa}{1-\kappa}H(\bm q_n^*; \sigma^2). 
\end{align*}
The second part of \Cref{thm:pert} follows from Bernstein's inequality combined with a union bound. Note that one needs to apply Bernstein's inequality twice due to the sample mean used in the sample variance estimator $\widehat{\sigma}^2$. 
\end{proof}

\section{Random subspaces}

\subsection{Proof of \Cref{thm:rb}}\label{app:mc1}
If we denote $\bar{f} = k^{-1}\sum_{i\in [k]}g_i\in\bar{V}_{n, k}$, then 
\begin{align*}
\E\left[\|\bar{f}-f\|^2_{L^2_\mu}\right] = \int_\Omega\E\left[|\bar{f}-f|^2\right] \mu(\d x) = \frac{1}{k}\|\sigma\|^2_{L^2_\mu}.
\end{align*}
By Markov's inequality, 
\begin{align}
&\P\left(\min_{v\in \bar{V}_{n,k}}\|f-v\|^2_{L^2_\mu}\geq\e^2\right)\leq\P\left(\|\bar{f}-f\|^2_{L^2_\mu}\geq\e^2\right)\leq \frac{\|\sigma\|^2_{L^2_\mu}}{k\e^2}\leq\frac{1}{2}&\text{($k = \lceil2\e^{-2}\|\sigma\|^2_{L^2_\mu}\rceil$)}.\label{markov}
\end{align}
Define $\bar{V}_{n, k}^i\coloneqq\left\{\sum_{j=(i-1)k+1}^{ik}\alpha_jg_j, \ \bm\alpha\in\R^k\right\}\subset \bar{V}_{n, k}$. It follows from a boosting argument that
{\small\begin{align*}
\P\left(\min_{v\in \bar{V}_{n,k}}\|f-v\|^2_{L^2_\mu}<\e^2\right)&\geq\P\left(\min_{i\in [l]}\min_{v\in \bar{V}_{n, k}^i}\|f-v\|^2_{L^2_\mu}<\e^2\right)= 1-\P\left(\min_{i\in [l]}\min_{v\in \bar{V}_{n, k}^i}\|f-v\|^2_{L^2_\mu}\geq\e^2\right)\\
& = 1- \prod_{i\in [l]}\P\left(\min_{v\in \bar{V}_{n, k}^i}\|f-v\|^2_{L^2_\mu}\geq\e^2\right)\stackrel{\eqref{markov}}{\geq} 1-\left(\frac{1}{2}\right)^l.
\end{align*}}
Choosing $l\geq \log(1/\delta)/\log 2$ and noting $2/\log 2<3$ yields the desired result.

\subsection{Proof of \Cref{thm:984}}\label{app:kernel}
Without loss of generality, we assume $c=1$; the general case can be considered by scaling. Let $s>2r+1$ and $n=2s$. For each random basis function $g_i = g(x; Z_i)\in V_{n}$, write $g_i = f+\sum_{j\in\N}\sqrt{\lambda_j}\xi_{ij}\phi_j$ where $\sqrt{\lambda_j}\xi_{ij}$ are the corresponding KL expansion coefficients as defined in \eqref{here}. 

For $i\in [s]$, we introduce the following (independent) centered functions as 
\begin{align*}
h_i(x) = g_{2i}(x) - g_{2i-1}(x) = \sum_{j\in\N}\sqrt{\lambda_j}\zeta_{ij}\phi_j(x) = h_{i, r}(x) + \bar{h}_{i,r}(x)\in V_n,
\end{align*}
where $\zeta_{ij} = \xi_{2i,j}-\xi_{2i-1, j}$ are mutually uncorrelated random variables with mean zero and variance $\E[\zeta_{ij}^2] = 2$, and $h_{i, r} = \sum_{j\in [r]}\sqrt{\lambda_j}\zeta_{ij}\phi_j$, $\bar{h}_{i, r} = \sum_{j>r}\sqrt{\lambda_j}\zeta_{ij}\phi_j$. Moreover, we let $\bm\xi_i = (\xi_{i1}, \ldots, \xi_{ir})^\top$, $\bm\zeta_i = (\zeta_{i1}, \ldots, \zeta_{ir})^\top$, and $\bm L = (\bm\zeta_1, \ldots, \bm\zeta_s)^\top\in\R^{s\times r}$.  

We now consider the intermediate approximant defined as $\widetilde{f}_n = g_n - \sum_{j\in [r]}\sqrt{\lambda_j}\xi_{nj}\phi_j$, which in general is not  an element in $V_n$. Under the non-atomic assumption, $\{\phi_i\}_{i\in [r]}$ and $\{h_{i, r}\}_{i\in [s]}$ span the same linear subspace a.s., which allows to represent $\{\phi_i\}_{i\in [r]}$ as a particular linear combination of $\{h_{i, r}\}_{i\in [s]}$ as follows:
\begin{align}
\widetilde{f}_n = g_n - \sum_{j\in [r]} \sqrt{\lambda_j}\xi_{nj}\phi_j = g_n -\sum_{i\in [s]}\theta_ih_{i, r}, \quad\quad\bm\theta = (\theta_1, \ldots, \theta_{s})^\top = (\bm L^\dagger)^\top\bm\xi_n. \label{ridge}
\end{align}

By Markov's inequality, it holds with probability at least $0.9$, 
\begin{align}
\left\|\widetilde{f}_n-f\right\|^2_{L^2_\mu} &= \left\|\sum_{j>r}\sqrt{\lambda_j}\xi_{nj}\phi_j\right\|^2_{L^2_\mu}\leq  10\E\left[ \left\|\sum_{j>r}\sqrt{\lambda_j}\xi_{nj}\phi_j\right\|^2_{L^2_\mu}\right]= 10\sum_{j>r}\E[\lambda_j\xi^2_{nj}]=10\tau_{r+1}.\label{hgf1}
\end{align}

To find a substitute of $\widetilde{f}_n$ in $V_n$, based on $\bm\theta$, we define $f_{n} = g_n-\sum_{i\in [s]}\theta_ih_{i}\in V_{n}$. 
By the Cauchy--Schwarz inequality, 
\begin{align}
\left\|\widetilde{f}_n-f_n\right\|^2_{L^2_\mu} = \left\|\sum_{i\in [s]}\theta_i\bar{h}_{i, r}\right\|^2_{L^2_\mu}\leq s\|\bm\theta\|_2^2\cdot\frac{1}{s}\sum_{i\in [s]}\|\bar{h}_{i,r}\|^2_{L^2_\mu}.\label{hgf2}
\end{align}

Applying Markov's inequality again, we obtain that with probability of at least $0.8$, 
\begin{align*}
\frac{1}{s}\sum_{i\in [s]}\|\bar{h}_{i,r}\|^2_{L^2_\mu}&\leq 10\E_{Z_1}\left[\|\bar{h}_{1,r}\|^2_{L^2_\mu}\right]= 20\tau_{r+1}\\
s\|\bm\theta\|_2^2&= s\|(\bm L^{\dagger})^\top\bm\xi_n\|_2^2\leq \frac{s\|\bm\xi_n\|_2^2}{\lambda_{\min}(\bm L^\top\bm L)}\leq\frac{10\E[\|\bm\xi_n\|_2^2]\cdot s}{\lambda_{\min}(\bm L^\top\bm L)}=\frac{10rs}{\lambda_{\min}(\bm L^\top\bm L)}, 
\end{align*}
where $\lambda_{\min}(\bm L^\top\bm L)$ denotes the smallest eigenvalue value of $\bm L^\top\bm L$. To further bound $\lambda_{\min}(\bm L^\top\bm L)$ from below, we use matrix concentration inequalities. Note that $\bm L^\top\bm L = \sum_{i\in [s]}\bm\zeta_{i}\bm\zeta_i^\top$ is a sum of i.i.d. rank-one matrices with $\E[\bm\zeta_{i}\bm\zeta_i^\top]=2\bm I$. A straightforward idea is to apply the Chernoff bound to obtain a lower bound for $\lambda_{\min}(\bm L^\top\bm L)$. However, since $\lambda_{\max}(\bm\zeta_i\bm\zeta_i^\top)$ is not uniformly bounded with probability one, a direct argument does not apply. To address this, we apply a truncation argument. 

Let $T=\max\left\{4\log(4r), 2\sqrt{\log(sr)}\right\}$ be the truncation parameter, and define $\widetilde{\zeta}_{ij}$ as follows:
\begin{align*}
\widetilde{\zeta}_{ij} = \begin{cases}
\zeta_{ij}& |\zeta_{ij}|\leq T\\
0& \text{else}
\end{cases}, 
\quad\quad\widetilde{\bm\zeta}_{i} = (\widetilde{\zeta}_{i1}, \ldots, \widetilde{\zeta}_{ir})^\top.
\end{align*}
Under the tail assumption \eqref{subexp}, it follows from a union bound estimate that, with probability at least $0.9$, for all $i\in [s]$, $\bm\zeta_{i} = \widetilde{\bm\zeta}_{i}$. 
Meanwhile, for $j, j'\in [r]$, if $j \neq j'$, applying the Cauchy--Schwarz inequality, 
\begin{align}
\left|\E[\widetilde{\zeta}_{ij}\widetilde{\zeta}_{ij'}] - \E[\zeta_{ij}\zeta_{ij'}]\right| &= \left|\E[\widetilde{\zeta}_{ij}(\widetilde{\zeta}_{ij'}-\zeta_{ij'})]\right|+\left|\E[(\widetilde{\zeta}_{ij}-\zeta_{ij})\zeta_{ij'}]\right|\nonumber\\
&\leq 2\left(\max_{i,j}\E[\zeta^2_{ij}]\right)^{\frac{1}{2}}\left(\max_{i, j}\E[(\widetilde{\zeta}_{ij}-\zeta_{ij})^2]\right)^{\frac{1}{2}}\nonumber\\
&\stackrel{\eqref{subexp}}{\leq}2\sqrt{2}\left(\max_{i, j}\E[(\widetilde{\zeta}_{ij}-\zeta_{ij})^2]\right)^{\frac{1}{2}}\nonumber\\
&=2\sqrt{2}\left(\max_{i, j}\E[\mathbf 1_{\{\zeta_{ij}>T\}}\zeta_{ij}^2]\right)^{\frac{1}{2}}\nonumber\\
&\stackrel{\eqref{subexp}}{\leq}4\sqrt{(T+1)e^{-T}}\nonumber\\
& \leq 4e^{-T/4}\leq\frac{1}{r}.\label{yonghuo}
\end{align}
A similar bound also holds for the case when $j=j'$. Applying Weyl's inequality, 
\begin{align*}
\lambda_{\min}(\E[\bm\zeta_{1}\bm\zeta^\top_{1}])-\lambda_{\min}(\E[\widetilde{\bm\zeta}_{1}\widetilde{\bm\zeta}^\top_{1}])\leq \|\E[\bm\zeta_{1}\bm\zeta^\top_{1}]-\E[\widetilde{\bm\zeta}_{1}\widetilde{\bm\zeta}^\top_{1}]\|_2\leq\|\E[\bm\zeta_{1}\bm\zeta^\top_{1}]-\E[\widetilde{\bm\zeta}_{1}\widetilde{\bm\zeta}^\top_{1}]\|_F\leq 1. 
\end{align*}
Consequently, $\lambda_{\min}(\E[\widetilde{\bm\zeta}_{1}\widetilde{\bm\zeta}^\top_{1}])\geq 1$. 
Since $\lambda_{\max}(\widetilde{\bm\zeta}_i\widetilde{\bm\zeta}_i^\top) = \|\widetilde{\bm\zeta}_i\|^2_2\leq rT^2$, by the matrix Chernoff bound \cite{tropp2012user}, 
\begin{align}
\P\left(\lambda_{\min}\left(\sum_{i\in [s]}\widetilde{\bm\zeta}_i\widetilde{\bm\zeta}_i^\top\right)< 0.5\cdot s\right)\leq r\cdot (0.9)^{\frac{s}{2rT^2}}\leq 0.5,\label{yonghuo2}
\end{align}
where the last step holds if choosing $s = c'r(\log r)^3$, where $c'>0$ is some sufficiently large absolute constant (independent of $r$). 
Combining \eqref{yonghuo} and \eqref{yonghuo2} yields that, with probability at least $0.4$, $\lambda_{\min}(\bm L^\top\bm L)\geq s/2$. 
This combined with \eqref{hgf1} and \eqref{hgf2} yields that, with probability at least $0.1$, 
\begin{align*}
\left\|f-f_n\right\|_{L^2_\mu}\leq \left\|\widetilde{f}_n-f\right\|_{L^2_\mu} + \left\|\widetilde{f}_n-f_n\right\|_{L^2_\mu}\leq\sqrt{10\tau_{r+1}} + \sqrt{400r\tau_{r+1}}\leq 24\sqrt{r\tau_{r+1}}. 
\end{align*}
The proof is finished by applying a similar boosting argument in \Cref{thm:rb} to lift the constant probability in both cases to $1-\delta$. 

\end{appendix}

\end{document}